\documentclass{article}
\usepackage{graphics,multirow,amsmath,amssymb,textcomp,subfigure,multirow,xspace,arydshln,cite}
\usepackage[]{graphicx}
\usepackage[comma]{natbib}
\usepackage{url}
\usepackage{amsmath}
\usepackage{amssymb}
\usepackage{algorithmic}
\usepackage{url}
\usepackage{caption}
\usepackage{tikz}
\usepackage{hyperref}
\usepackage{xcolor}
\usepackage{empheq}
\usepackage{fancybox}
\usepackage{moresize}
\usepackage[colorinlistoftodos]{todonotes}
\usepackage[top=1in, bottom=1.25in, left=1.25in, right=1.25in]{geometry}

\newcommand{\ltriplebar}{\lvert\kern-0.9pt\lvert\kern-0.9pt\lvert}
\newcommand{\rtriplebar}{\rvert\kern-0.9pt\rvert\kern-0.9pt\rvert}

\newcommand{\by}{\mathbf{y}}

\definecolor{PlotColorA}{HTML}{1f77b4}
\definecolor{PlotColorB}{HTML}{ff7f0e}
\definecolor{PlotColorC}{HTML}{2ca02c}
\definecolor{PlotColorD}{HTML}{d62728}
\definecolor{PlotColorE}{HTML}{9467bd}
\definecolor{PlotColorF}{HTML}{8c564b}
\definecolor{PlotColorG}{HTML}{e377c2}

\usepackage{soul}
\definecolor{HighlightColor}{HTML}{baffb2}
\sethlcolor{HighlightColor}

\newcommand{\tinyb}[1]{\scalebox{0.6}{{\normalsize #1}}}

\begin{document}

\title{Mobile Computational Photography: A Tour}

\author{Mauricio Delbracio$^1$, Damien Kelly$^1$, Michael S. Brown$^{2}$, Peyman Milanfar$^1$ \\[.5em]
$^1$Google Research, Mountain View, CA, USA \\
 \{mdelbra,damienkelly,milanfar\}@google.com \\[.2em]
$^2$ EECS Department, York University, Toronto, Canada \\ 
mbrown@eecs.yorku.ca}

\date{}
\maketitle

\begin{abstract}
The first mobile camera phone was sold only $20$ years ago, when taking pictures with one's phone was an oddity, and sharing pictures online was unheard of. Today, the smartphone is more camera than phone. How did this happen? This transformation was enabled by advances in computational photography---the science and engineering of making great images from small form factor, mobile cameras. Modern algorithmic and computing advances, including machine learning, have changed the rules of photography, bringing to it new modes of capture, post-processing, storage, and sharing. In this paper, we give a brief history of mobile computational photography and describe some of the key technological components, including burst photography, noise reduction, and super-resolution. At each step, we may draw naive parallels to the human visual system. 
\end{abstract}

\tableofcontents

\section{Introduction and historical overview}
\begin{figure}[h]
\centering
\includegraphics[width=5in]{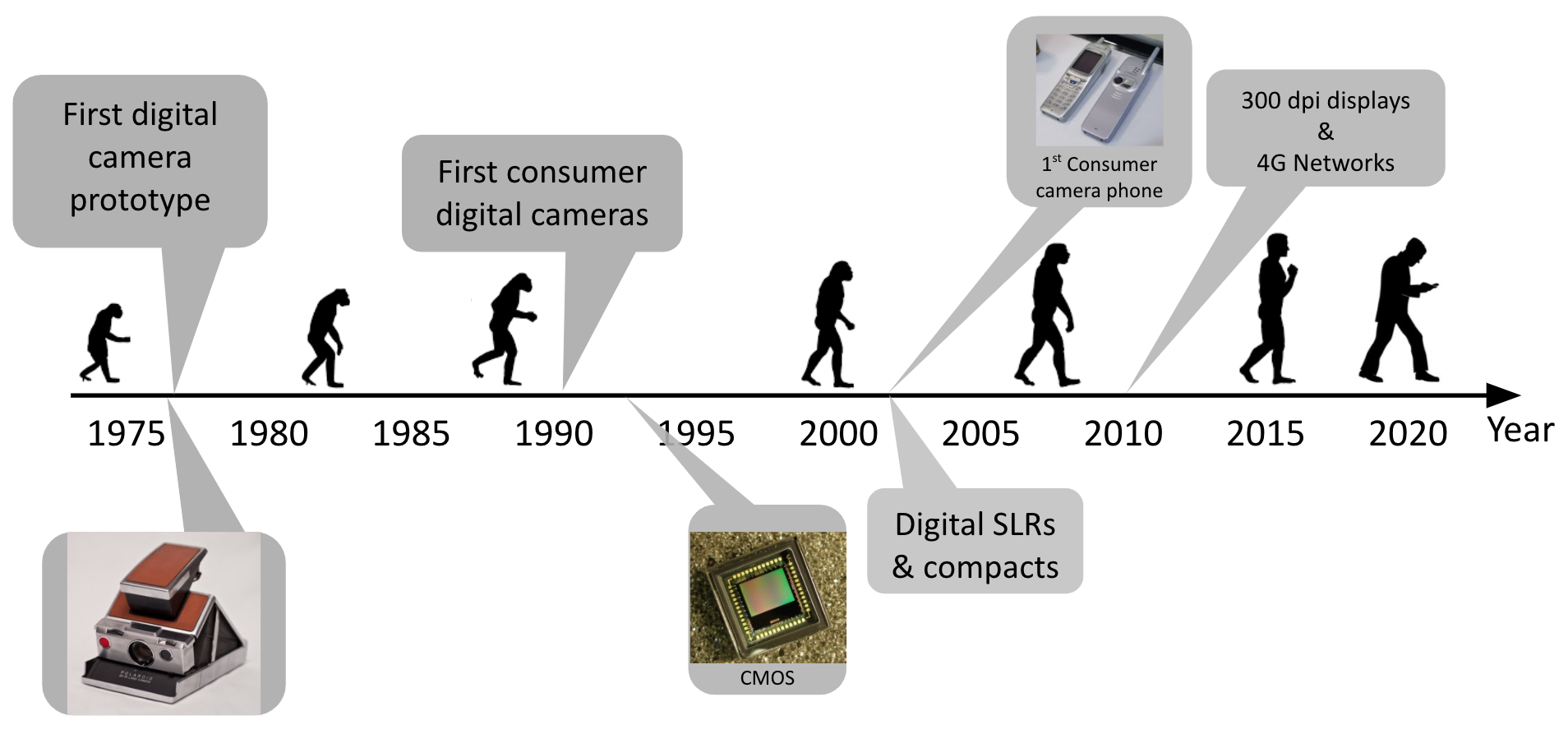}
\caption{An {\it incomplete} timeline of photographic innovations in the last four decades.}
\label{fig:history}
\end{figure}

\begin{figure}[h]
\centering
\includegraphics[width=5in]{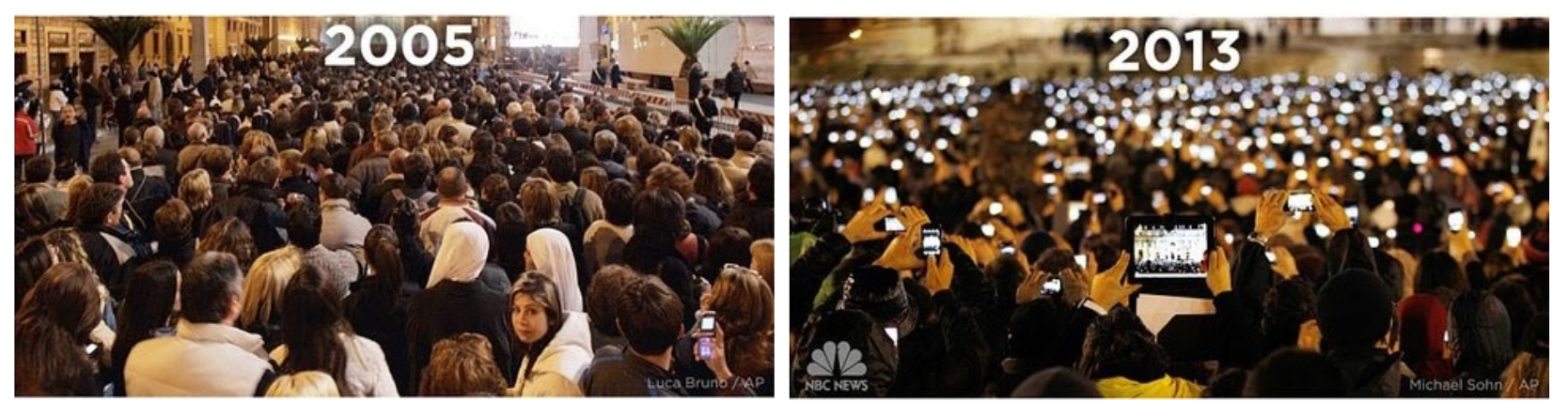}
\caption{Note the crowds at Saint Peter’s Square in Vatican City when Pope Benedict XVI was announced in 2005 and when Pope Francis was announced in 2013. Image courtesy of NBC News.}
\label{fig:vatican}
\end{figure}

Modern digital photography has a fascinating and nuanced history (Figure \ref{fig:history}), punctuated by important advances in both camera sensor technology and algorithms that operate on the captured signals. In this review, we will concentrate on the more recent two decades of intense and rapid progress in \emph{computational photography}. Even more specifically, we will provide an excursion through \emph{mobile} computational photography, which is where we've seen its largest impact on the daily lives of people across the globe. From news to social media, digital photographs (now overwhelmingly captured on mobile devices) have fundamentally transformed how we capture and remember our world. Indeed, it is not an exaggeration to say that the smartphone (and its camera in particular) has changed the world (see Figure~\ref{fig:vatican}). 

The era of analog (film) photography saw its golden age from the 1930s onward, when giants of the craft, like Ansel Adams, practiced their art of ``Making a Photograph''~\citep{AnselAdams} by hand in their custom dark rooms. Remarkably, innovations by Adams and others, such as the \emph{dodge and burn} technique for high-dynamic-range photography, have persisted in digital and computational photography to this day, albeit in much more formal, algorithmic incarnations. Analog film thus dominated the scene for nearly $50$ years but was largely discontinued in $2005$, when standalone point-and-shoot digital cameras became dominant, and before cellphones had good imaging capabilities. 

The first commercial digital cameras appeared in the early 1990s. In 1992, the first digital single-lens reflex (DSLR) cameras entered the market, but were prohibitively expensive, costing upwards of $\$20,000$. Unsurprisingly, these early devices failed to capture a large market share. The introduction of CMOS (complementary metal oxide semiconductor) image sensors in 1993 facilitated the development of what became known as “camera on a chip.” This revolutionary sensor would enable far less expensive devices to be built with proportionally better power efficiency. Yet, the technical difficulties in replacing the existing standard of the time (CCD arrays) were significant. These included noisier pixels and a rolling (rather than global) shutter in the cheaper CMOS sensors. Some of these technical difficulties meant that it would take another 10 years before CMOS systems would enable mass production of mobile and digital cameras. %

In the early 2000s, the dominant digital devices for photography were single-lens reflex (SLR) and digital point-and-shoot cameras. Twenty years hence, the prices of these two camera types are roughly in the same proportion, while the quality of each has significantly improved---meaning that both technologies now present a much better value for the user. In standalone cameras, somewhat surprisingly, this improvement is largely due to better sensors and optics, and to a much lesser degree due to technical advances in software. Even today, standalone cameras still have relatively naive software pipelines. Meanwhile, algorithmic and software innovations have thrived in mobile devices. Why this is the case is an interesting question. While many factors have contributed to this dichotomy---hardware advances on the standalone cameras vs. software advances on mobile platforms enabled by significant computing power---one thing remains clear: the requirements imposed by the sleek industrial design and consequent form factors in mobile smartphones have severely limited what can be done with imaging hardware. As a result, the best possible hardware on mobile devices is still often bettered by complementary development and close integration of algorithmic software solutions. In some ways, the situation is not altogether dissimilar to the development of vision in nature. The evolution of the physical form of the eye has had to contend with physical constraints that limit the size, shape, and sensitivity of light gathering in our vision system. In tandem and complementary fashion, the visual cortex has developed the \emph{computational} machinery to interpolate, interpret, and expand the limits imposed by the physical shape of our eyes---our visual ``hardware.''

The introduction of the iPhone in 2007 was a watershed moment in the evolution of mobile devices and changed the course of both phone and camera technology. Looking back at the early devices, the layperson may conclude that the camera was immediately transformed to become the high-utility application we know today. This was in fact not the case. What was reinvented for the better at the time were the display and user interface, but not necessarily the camera yet; indeed, the two-megapixel camera on the first iPhone was far inferior to just about any existing point-and-shoot camera of comparable price at the time. 

\begin{figure}[h]
\centering
\includegraphics[width=5in]{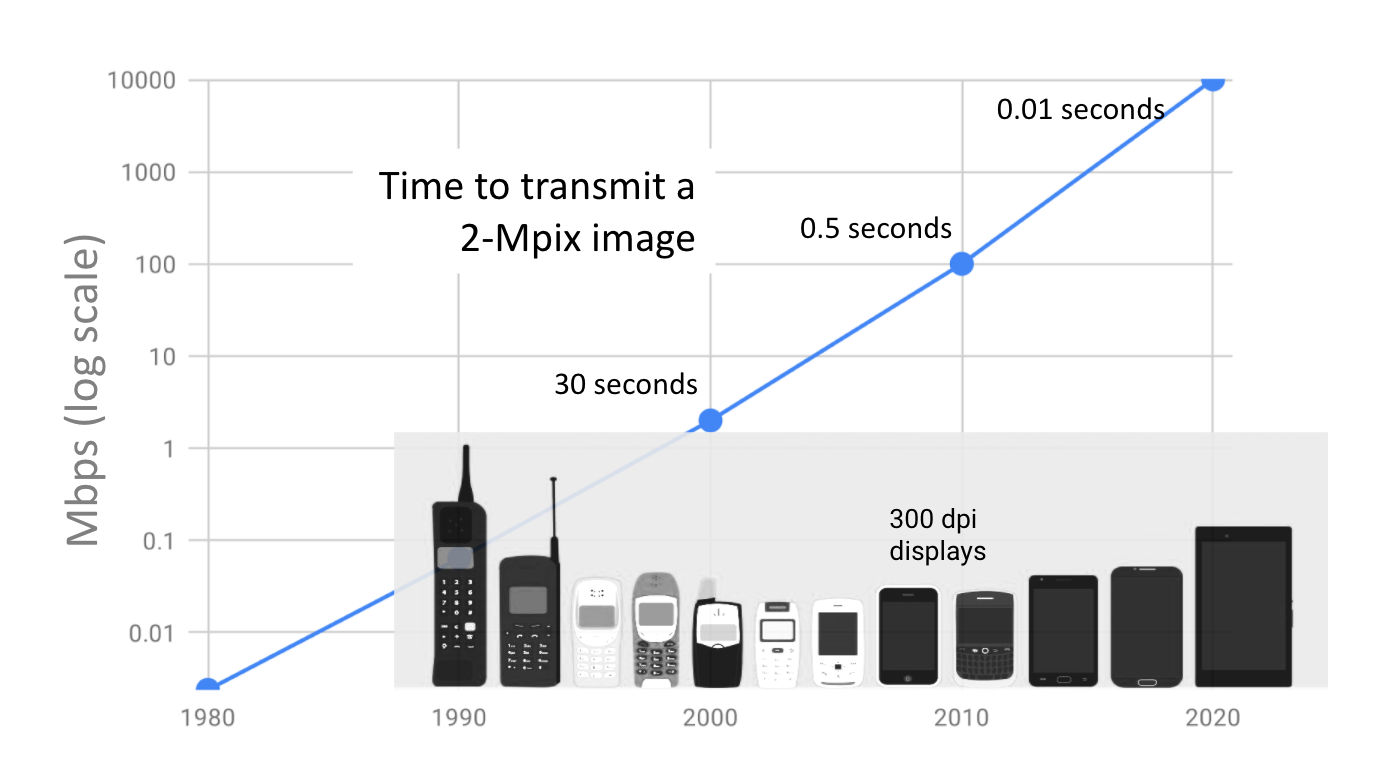}
\caption{The year 2010 saw the convergence of two important trends: improved displays and increased wireless speed. These forces conspired to catapult mobile photography to the dominant mode of imaging in the ensuing decade.}
\label{fig:DisplayWireless}
\end{figure}

The year 2010 was pivotal for the mobile camera. A transition to both 4G wireless and 300 dots per inch (dpi) displays enabled users to finally appreciate their photographs on the screens of their own mobile devices. Indeed, users felt that their phone screens were not only rich enough for the consumption of their own photos, but also sufficient to make photo-sharing worthwhile. Meanwhile, the significantly faster wireless network speeds meant that individuals could share (almost instantaneously) their photos (see Figure \ref{fig:DisplayWireless}). Once viewing and sharing had been improved by leaps and bounds, it became imperative for mobile manufacturers to significantly improve the quality of the captured photos as well. Hence began an emphasis on improved light collection, better dynamic range, and higher resolution for camera phones so that consumers could use their phones as cameras \emph{and} communication devices. The added bonus was that users would no longer have to carry both a phone and a camera---they could rely on their smartphone as a multipurpose device. 

To achieve this ambitious goal meant overcoming a large number of limitations--challenges that have kept the computational photography community busy for the last decade. In the next section, we begin by highlighting some of the main limitations of the mobile imaging platform as compared to standalone devices, and go on in the rest of the paper to review how these challenges have been met in research and practice. 

\begin{figure}
\centering
\includegraphics[width=3in]{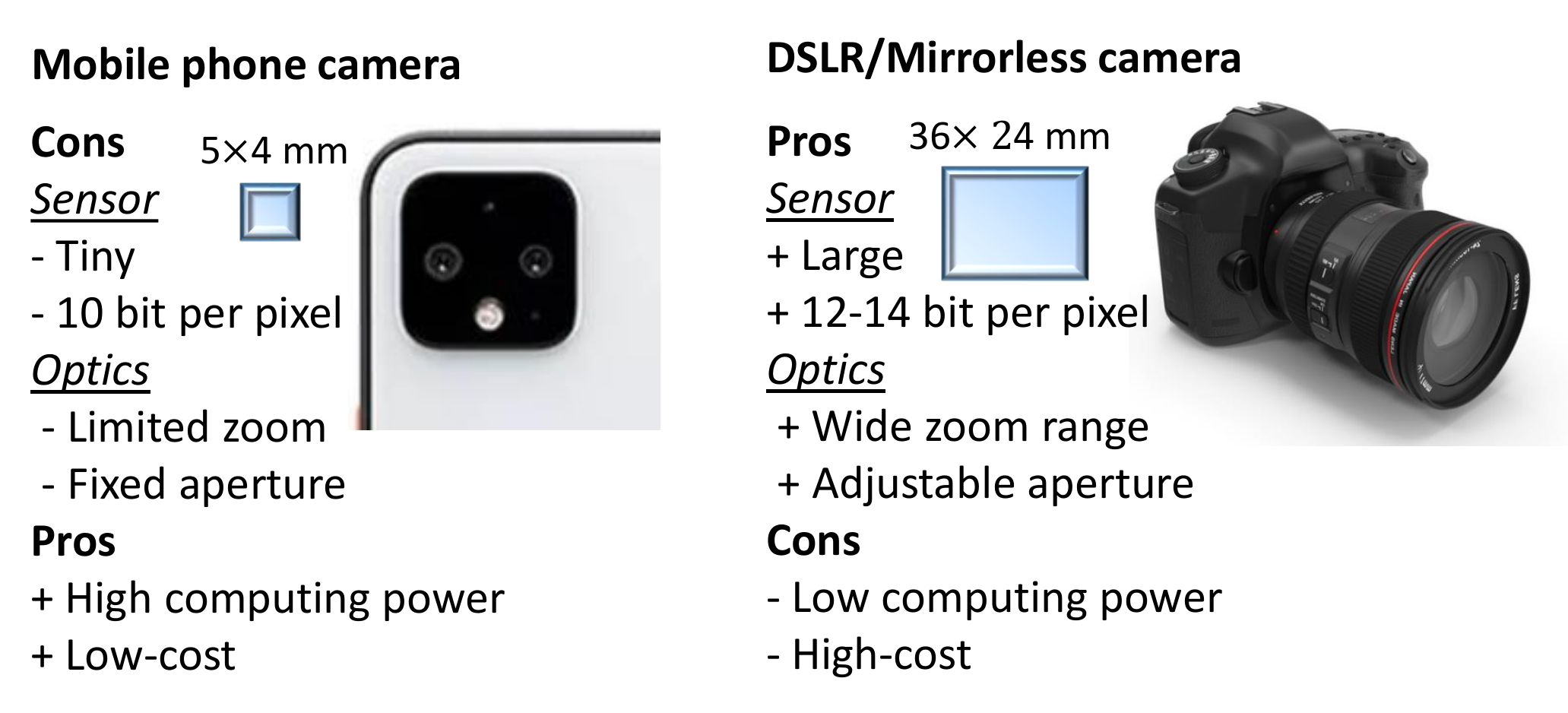}
\caption{This figure shows the pros and cons of a smartphone compared to a DSLR. The most notable differences are the larger sensor and optics available on a DSLR.  Surprisingly, however, a high-end smartphone has significantly more computing power than most DSLR cameras.}
\label{ProCons}
\end{figure}

\section{The Mobile Camera: Hardware and its Limitations}\label{sec:hardware_and_limitations}

Ideally, a smartphone camera would offer photographic performance on par with a DSLR (or at least a compact) camera.  However, the smartphone camera has several notable disadvantages due to constraints placed on its form factor in order for it to be integrated in the phone's thin profile.  Figure~\ref{ProCons} shows a smartphone with integrated cameras next to a typical DSLR camera.  The physical camera sensor and associated lens optics of the smartphone camera are significantly smaller and less flexible than those on a DSLR camera.  Yet, while the smartphone's physical hardware is limited, the smartphones have access to much more computing power than available on a DSLR. To draw a rough, but stark contrast between the two platforms, a mobile camera's small aperture limits light collection by two orders of magnitude as compared to a typical DSLR. Meanwhile, the same mobile device houses roughly two orders of magnitude more computing power. The trade-off of additional computing for more sophisticated imaging hardware is thus inevitable.

Next, we briefly summarize several key limitations of the mobile phone camera as compared to a DSLR. 

\subsection{Sensor size and limited aperture}

The most obvious limitations of a smartphone camera are the size of its sensor and compactness of its optics.  Modern smartphone sensors are roughly 5$\times$4 mm in size, while many DSLR cameras still use full-size 36$\times$24 mm sensors. In addition to a small sensor size, the optics of the mobile phone camera are significantly smaller and less adjustable compared to a typical lens used on a DSLR. In addition, most mobile cameras use a compact lens array that has a fixed aperture. The focal length is also limited, leading many phone makers to have two or more cameras with different focal lengths, each serving a different purpose (main, zoom, wide, etc.) 

\subsection{Noise and limited dynamic range}
Image noise can be defined as random unwanted variations of the intensity level on image pixels. In addition to the random fluctuations due to thermal agitation in electronics, there exists a permanent, unavoidable source of noise due to the discrete nature of light (photon shot noise).

With the smaller aperture and sensor size, for a given exposure time, a smartphone is able to have at best a small fraction of the amount of light that would be captured by a DSLR. A smaller sensor also means that even less light hits the surface of the sensor when capturing an image.  As a result, smartphone cameras often need to apply a non-trivial multiplicative gain to the recorded signal.  This gain is controlled by the ISO setting---a higher ISO number implies an increase in the gain factor, which amplifies the sensor noise. Consequently, the smartphone camera images produced at the sensor are markedly more noisy than images captured with DSLR sensors.  

Another notable difference between DSLR and smartphone cameras is the dynamic
range of the sensor, which is defined as the ratio between the full well capacity of a pixel's photodiode at maximum gain, and its noise (read noise). In practice, this defines the brightest and darkest parts of the scene that can be captured without clipping or saturation. The dynamic range is directly correlated to the pixel size. A DSLR pixel’s photodiode is roughly 4 microns in width, while a smartphone sensor is closer to 1.5 microns or less. This means that pixels of a smartphone sensor have a much smaller well capacity and therefore the maximum amount of electrical current they can capture at each photodiode is reduced. As a result, a DSLR can effectively encode anywhere between 4096 (12 bits) and 16384 (14 bits) tones per pixel. Whereas a typical smartphone camera sensor is limited to 1024 (10 bits) of tonal values per pixel.

\subsection{Limited depth of field}

The depth of field (DoF) defines the region in the image of the scene where objects appear sharp.  The DoF can be controlled by the camera's focal length and aperture.  The wider the aperture, the shallower the DoF.  In photography, especially when imaging human subjects for portraits, it is often desirable to have a narrow DoF to focus on the subject's face while blurring out the background.   The small aperture used on a mobile camera exhibits little DoF blur.  In addition, smartphone cameras have fixed apertures that do not allow for DoF adjustment at capture time. To overcome this limitation, most smartphones now provide a {\it synthetic} depth of field blur referred to as digital bokeh (see Section~\ref{sec:synthetic_bokeh}).

\subsection{Limited zoom}
As noted earlier, in response to consumer demands, smartphone design has trended towards ultra-thin form factors. This design trend imposes severe limitations on the thickness (or \emph{z-height}) of the smartphone’s camera module, limiting the effective focal length, which in turn limits the camera module’s optical zoom capability. To overcome this \emph{z-height} limitation, modern smartphone manufacturers typically feature multiple camera modules with different effective focal lengths and field of view, enabling zoom capabilities ranging from ultra-wide to telephoto zoom. The \emph{z-height} form factor restriction has spurred a so-called \emph{thinnovation} (a portmanteau on thin and innovation) in optical design, with manufacturers exploring folded optics in an effort to increase the optical path and effective focal length beyond the physical \emph{z-height} limits of the device.

\subsection{Color sub-sampling}
\begin{figure}[h]
\centering
\includegraphics[width=5in]{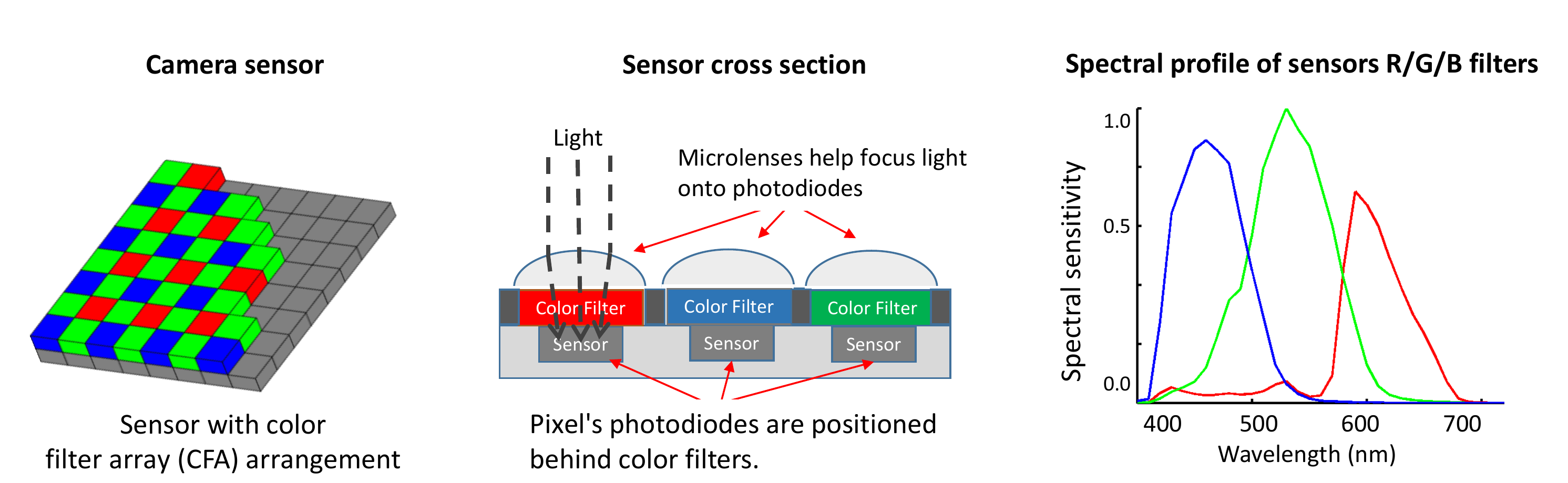}
\caption{A typical camera sensor with a color filter array layout (Bayer pattern) is shown.  A cross section of the sensor is shown along with an example of the spectral sensitivities of the color filters.}
\label{fig:oneshot}
\end{figure}

Finally, a key limitation for both smartphones and most DSLRs is that the sensors have only a single color filter associated with each pixel's photodiode\footnote{~Exceptions do exist, including sensors developed by Foveon and others, though these are not in common use.}, shown in Figure~\ref{fig:oneshot}.  This is analogous to how the human eye's cone cells are categorized by their sensitivity to either short-wavelength, medium-wavelength, or long-wavelength light.  Of course for any camera, the ultimate goal is three color values per pixel.  As a result, an interpolation process (called \emph{demoasicing}) is required to convert the sensor's sub-sampled color image to one having a three-channel (red, green, and blue; RGB) value at each pixel.  In addition, the   RGB color filters used on the camera sensor do not correspond to the perceptual-based CIE XYZ matching functions~\citep{Jiang:ColorFilterArrays2013}.  As a result, the ability to produce correct colorimetric measurements is often limited, and related to the color filters used.  

\section{The Camera Imaging Pipeline}

In this section we provide an overview of the steps applied by a digital camera to process the image recorded by the sensor and produce the final image that represents a high-quality, visually pleasing ``photo.''  These processing steps are applied in sequence and thus form a {\it pipeline} where the image pixel values are transformed step by step to produce the final image.  Camera systems have a dedicated chip, referred to as an image signal processor (ISP), which performs this processing pipeline in a matter of milliseconds for each image.

We first describe the basic sensor design, followed by a description of a typical single-image camera pipeline.   Single-image pipelines are still common on DSLR devices and sometimes used by mobile phone cameras under good lighting conditions.  We then describe multi-frame (or burst mode) pipelines that capture and fuse multiple images per photo, more typically used in mobile cameras.  Recent advances in multi-frame imaging have been instrumental in helping mobile phone cameras overcome many of the limitations described in the previous section. 

\subsection{Camera sensor}\label{sec:camera_sensor}

A camera sensor is comprised of a 2D grid of photodiodes.  A photodiode is a semiconductor device that converts photons (light radiation) into electrical charge.  A single photodiode typically corresponds to a single image pixel.  In order to produce a color image, color filters are placed over the photodiodes.  These color filters roughly correspond to the long, medium, and short cone cells found in the retina.  The typical arrangement of this color filter array (CFA) is often called a Bayer pattern, named after Bryce Bayer, who proposed this design at Kodak in 1975~\citep{Bayer:Patent}.   The CFA appears as a mosaic of color tiles laid on top of the sensor as shown in Figure \ref{fig:oneshot}. A key process in the camera pipeline is to ``demosaic'' the CFA array by interpolating a red, green, and blue value for each pixel based on the surrounding R, G, B colors.  It is important to note that the spectral sensitivities of the red, green, and blue color filters are {\it specific} to a particular sensor's make and model.  Because of this, a crucial step in the camera imaging pipeline is to convert these sensor-specific RGB values to a device-independent perceptual color space, such as CIE 1931 XYZ.  An image captured directly from a sensor that is still in its mosaiced format is called a {\it Bayer image} or {\it Bayer frame}.

\subsection{The camera pipeline}\label{sec:single_frame_pipeleine}

\begin{figure}[h]
\centering
\includegraphics[width=5in]{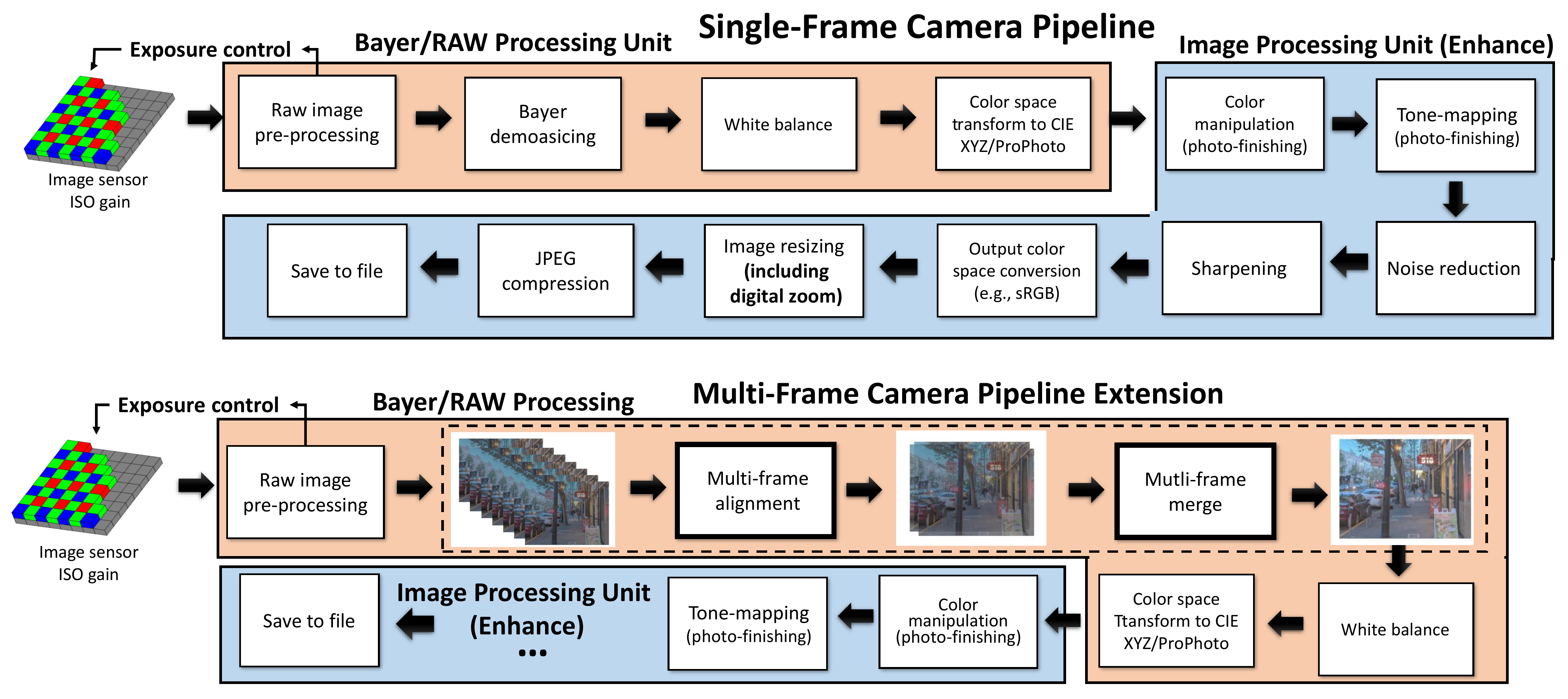}
\caption{The top of this figure shows a standard single-frame camera pipeline.  The bottom figure shows the extension to multi-frame (or burst imaging) used by most modern smartphone cameras.}
\label{fig:singleandmultipipeline}
\end{figure}

Figure~\ref{fig:singleandmultipipeline} (top) shows a diagram of a typical camera imaging pipeline that would be implemented by an ISP. Depending on the ISP design, the routines shown may appear in a slightly different order.  Many of the routines described would represent proprietary algorithms specific to a particular ISP manufacturer.  Two different camera manufacturers may use the same ISP hardware, but can tune and modify the ISP's parameters and algorithms to produce images with a photographic quality unique to their respective devices.  
The following provides a description of each of the processing steps outlined in Figure~\ref{fig:singleandmultipipeline} (top).

\paragraph*{Sensor frame acquisition:}~~When the Bayer image from the camera's sensor is captured and passed to the ISP, the ISO gain factor is adjusted at capture time depending on the scene brightness, desired shutter speed, and aperture.  The sensor Bayer frame is considered an unprocessed image and is commonly referred to as a {\it raw} image.  As shown in Figure~\ref{fig:oneshot}, the Bayer frame has a single R, G, B value per pixel location. These raw R, G, B values are not in a perceptual color space but are specific the to color filter array's spectral sensitivities.

\begin{figure}[h]
\centering
\includegraphics[width=5in]{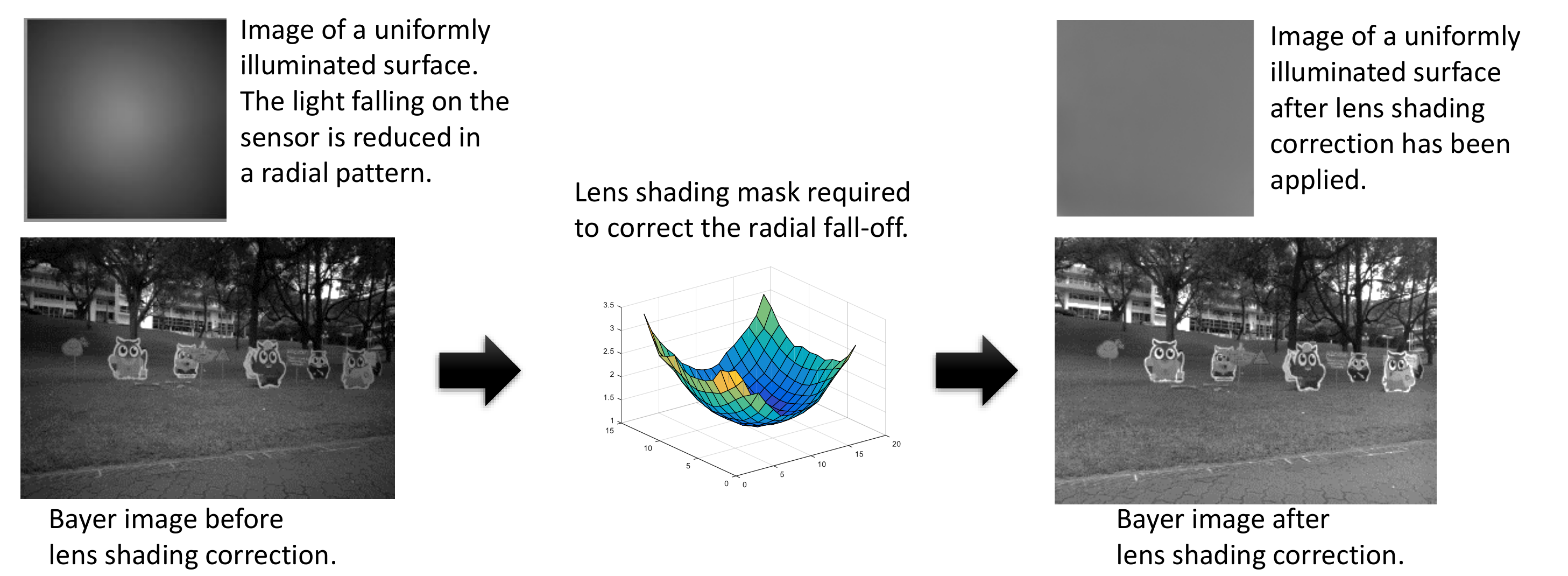}
\caption{Light entering the camera does not fall evenly across the sensor.  This creates an undesired vignetting effect.  Lens shading correction is used to adjust the recorded values on the sensor to have a uniform response.}
\label{fig:flatfield}
\end{figure}

\paragraph*{Raw-image pre-processing:}~~ The raw sensor image is normalized such that its values range from 0 to 1.  Many cameras provide a \textit{BlackLevel} parameter that represents the lowest pixel value produced by the sensor.  Interestingly, this deviates from $0$ due to sensor error. For example, a sensor that is exposed to no light should report a value of $0$ for its output, but instead outputs a small positive value called the BlackLevel.  This BlackLevel is subtracted off the raw image.  The BlackLevel is often image specific and related to other camera settings, including ISO and gain.  An additional \textit{WhiteLevel} (maximum value) can also be specified.  If nothing is provided, the \texttt{min} and \texttt{max} value of all intensities in the image is used to normalize the image between 0 and 1 after the BlackLevel adjustment has been applied.

The pre-processing stage also corrects any defective pixels on the sensor. A {\it defect pixel mask} is pre-calibrated in the factory and marks locations that have malfunctioning photodiodes.  Defective pixels can be photodiodes that always report a high value (a hot pixel) or pixels that output no value (a dead pixel).  Defective pixel values are interpolated using their neighbors.

Finally, a lens shading (or flat field) correction is applied to correct the effects of uneven light hitting the sensor.  The role of lens shading correction is shown in Figure~\ref{fig:flatfield}.  The figure shows the result of capturing a flat illumination field before lens shading correction.  The amount of light hitting the sensor falls off radially towards the edges.  The necessary radial correction is represented as a lens shading correction mask that is applied by the ISP to correct the effects from the non-uniform fallout. The lens shading mask is pre-calibrated by the manufacturer and is adjusted slightly per frame to accommodate different brightness levels, gain factors, and the estimated scene illumination used for white-balance (described below).   

\begin{figure}[h]
\includegraphics[width=5in]{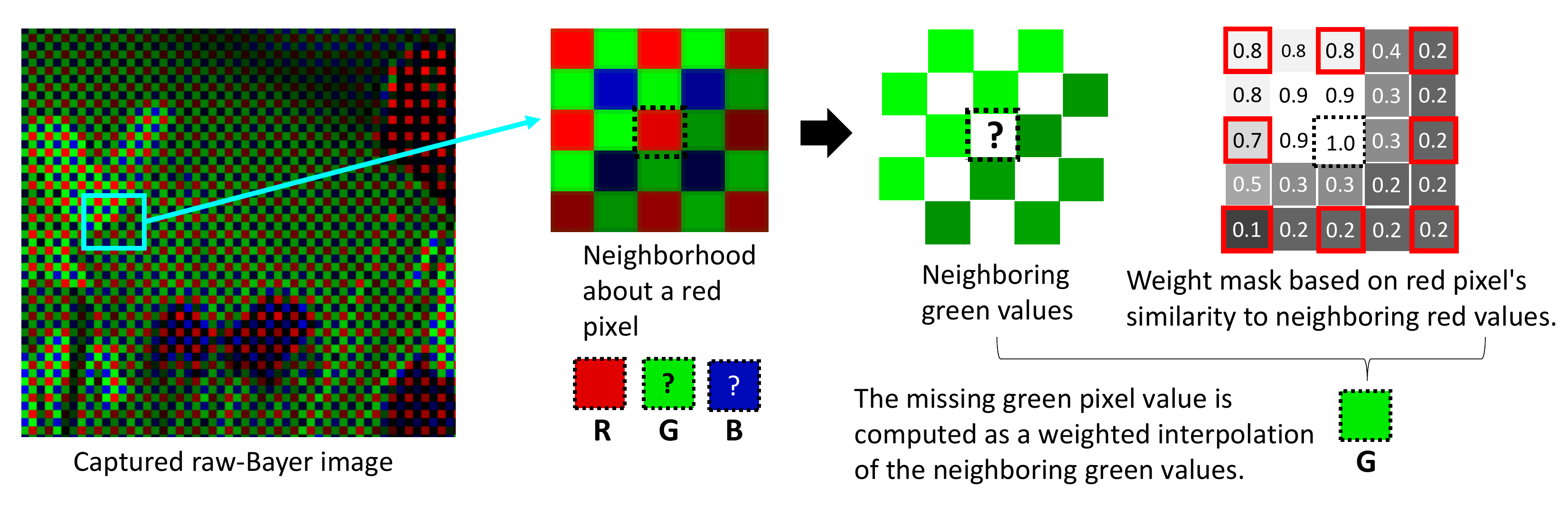}
\centering
\caption{This figure illustrates a common approach to image demosiacing.  Shown is a red pixel and its neighboring Bayer pixels.
The missing green and blue pixel values need to be estimated. These
missing values are interpolated from the neighboring pixels.  A weight mask based on the red pixel's similarity to its neighbors is computed to guide this interpolation. This weighted interpolation helps to avoid blurring across scene edges. This figure shows the interpolation of the missing green pixel value.}
\label{fig:demosaic}
\end{figure}

\paragraph*{Bayer demosaicing:}~~A demosaicing algorithm is applied to convert the single channel raw image to a three-channel full-size RGB image.  Demosaicing is performed by interpolating the missing values in the Bayer pattern based on neighboring values in the CFA. 
Figure~\ref{fig:demosaic} shows an example of the demosaicing process.  In this example, a zoomed photodiode with a red color filter is shown.  This pixel's green and blue color values need to be estimated.  These missing pixel values are estimated by interpolating the missing green pixel using the neighboring green values.  A per-pixel weight mask is computed based on the red pixel's similarity to neighboring red pixels. The use of this weight mask in the interpolation helps to avoid blurring around scene edges. Figure~\ref{fig:demosaic} illustrates a simplistic and generic approach, whereas most demosaicing algorithms are proprietary methods that often also perform highlight clipping, sharpening, and some initial denoising~\citep{Longere:DemosaicingAlgorithms}\footnote{~The astute reader will note that this demosaicing step is effectively interpolating two out of three colors at {\em every} pixel in the output image. The naive consumer may be shocked to learn that $2/3$ of their image is made up!}.

\begin{figure}[h]
\centering
\includegraphics[width=5in]{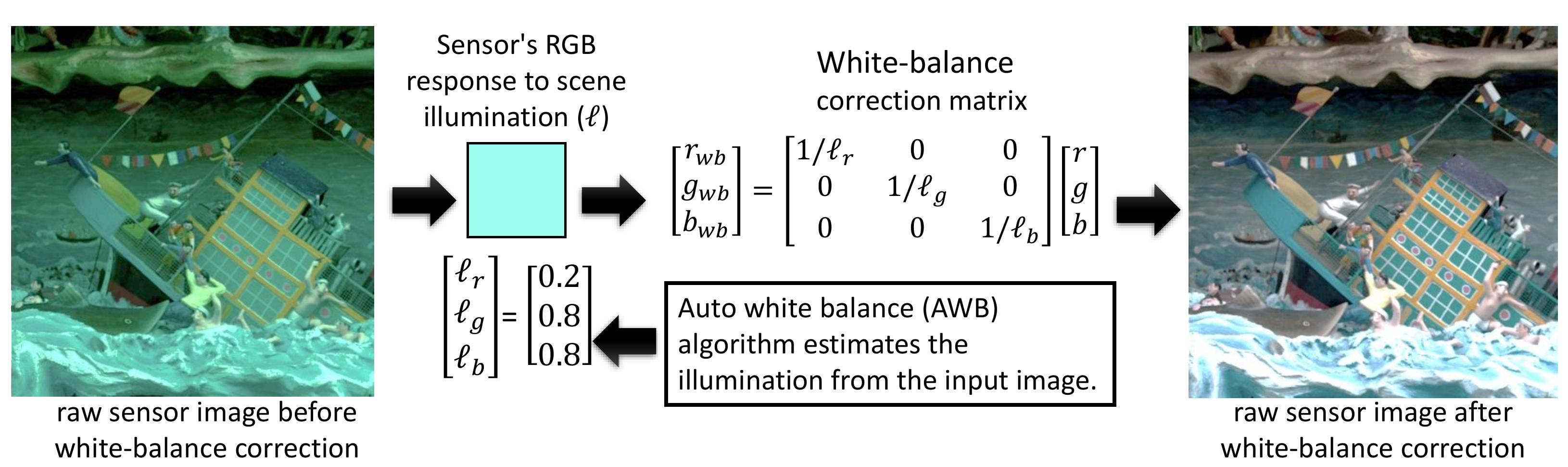}
\caption{White balance is applied to the image to mimic our visual system's ability to perform color constancy.  An auto white balance (AWB) algorithm estimates the sensor's response to the scene illumination.  The raw RGB values of the image are then scaled to based on the estimated illumination.}
\label{fig:whitebalance}
\end{figure}

\paragraph*{White Balance:}~~White balance is performed to mimic the human visual system's ability to perform chromatic adaptation to the scene illumination.  White balance is often referred to as computational color constancy to denote the connection to the human visual system.  White balance requires an estimate of the sensor's R, G, B color filter response to the scene illumination.  This response can be pre-calibrated in the factory by recording the sensor's response to spectra of common illuminations (e.g., sunlight, incandescent, and fluorescent lighting).  These pre-calibrated settings are then part of the camera's white-balance preset that a user can select.  A more common alternative is to rely on the camera's auto-white-balance (AWB) algorithm that estimates the sensor's R, G, B response to the illumination directly from the captured image.  Illumination estimation is a well-studied topic in computer vision and image processing with a wide range of solutions~\citep{AWB:GE:TIP07, AWB:GW:JFI80, AWB:GEHLER:CVPR08, NUS:JOSA:2014, AWB:Effective:CVPR15, AWB:FC4:CVPR17, AWB:FF:CVPR17}.  Figure~\ref{fig:whitebalance} illustrates the white-balance procedure.

Once the sensor's R, G, B values of the scene illumination have been obtained either by a preset or by the AWB feature, the image is modified (i.e., white-balanced) by dividing all pixels for each color channel by its corresponding R, G, B illumination value. This is similar to the well-known diagonal von Kries color adaption transform~\citep{VonKries}.  The Von Kries model is based on the response of the eye's short, medium, and long cone cells while white balance uses the sensor's R, G, B color filter responses.  

\paragraph*{Color space transform:}~~After white balance is applied, the image is still in the sensor-specific RGB color space.   The color space transform step is performed to convert the image from the sensor's raw-RGB color space to a device-independent perceptual color space derived directly from the CIE 1931 XYZ color space.   Most cameras use the wide-gamut ProPhoto RGB color space~\citep{susstrunk1999standard}.  ProPhoto is able to represent 90\% of colors  visible to the average human observer.  

\begin{figure}[h]
\centering
\includegraphics[width=5in]{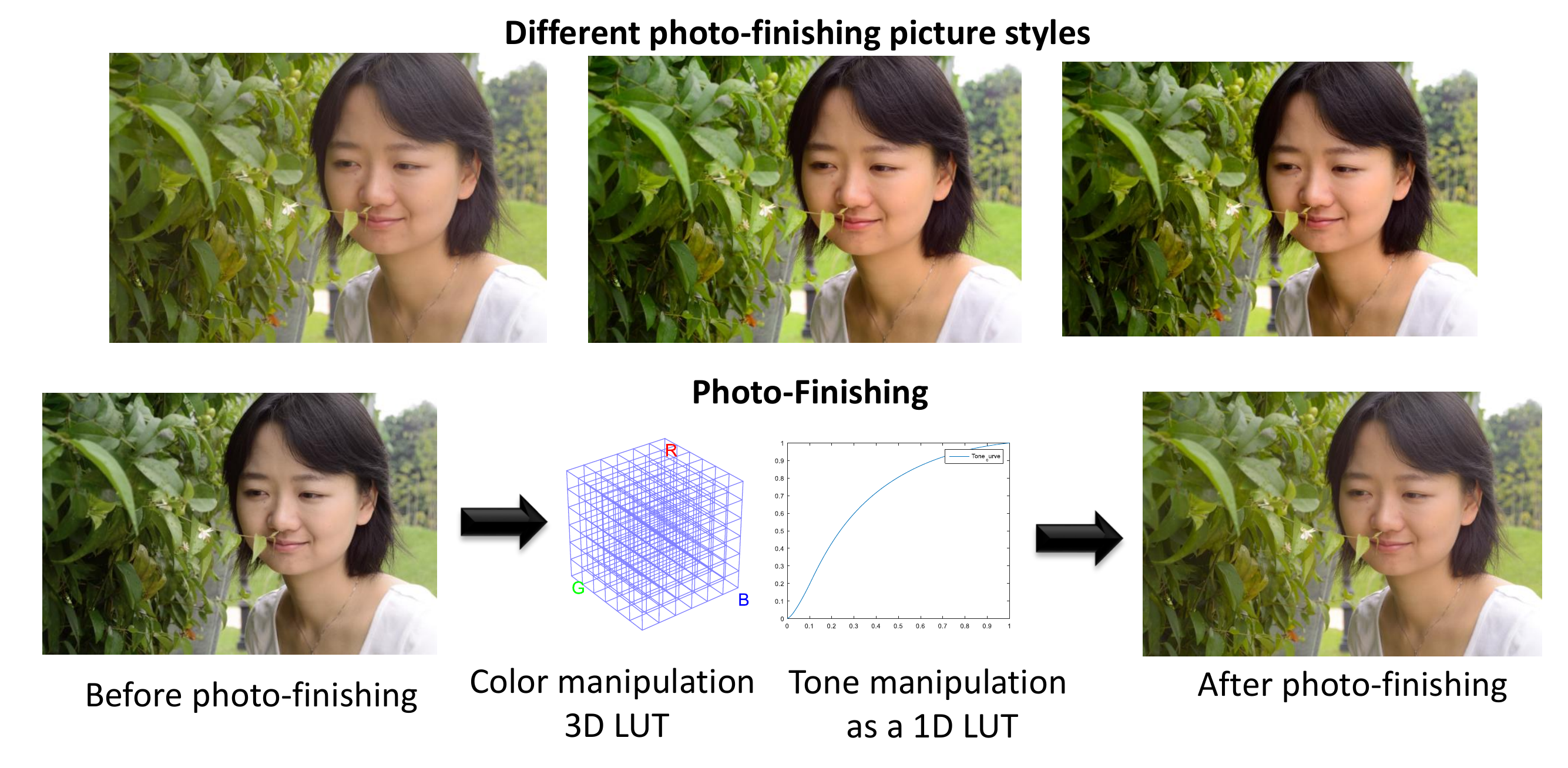}
\caption{Photo-finishing is used to enhance the aesthetic 
quality of an image.  Cameras often have multiple picture styles.
The color manipulation is often performed as a combination of a 
3D lookup table to modify the RGB colors and a 1D lookup table
to adjust the image's tonal values.}
\label{fig:photofinishing}
\end{figure}

\paragraph*{Color manipulation:}~~ Once the image is in a perceptual color space, cameras apply proprietary color manipulation to enhance the visual aesthetics of the image.  For DSLR devices, this enhancement can be linked to different picture styles or photo-finishing modes that the user can select, such as {\em vivid}, {\em landscape}, {\em portrait}, and {\em standard}.  Such color  manipulation is often implemented as a 3D lookup table (LUT) that is used to map the input ProPhoto RGB values to new RGB values based on a desired manipulation.  Figure~\ref{fig:photofinishing} shows an example. A 1D LUT tone map (discussed next) is also part of this photo-finishing manipulation.

Additional color manipulation may be performed on a smaller set of select colors used to enhance skin tones.  Establishing the 3D LUT can be a time-consuming process and is often performed by a group of ``golden eye'' experts who tune the ISP algorithms and tables to produce a particular photographic aesthetic often associated with a particular camera.   Note that camera manufacturers may even sell the same camera with different color manipulation parameters based on the preferences of users in different geographical locations.  For example, cameras sold in Asia and South America often have a slightly more vivid look than those sold in European and North American markets.

\paragraph*{Tone mapping:}~~A tone map is a 1D LUT that is applied per color channel to adjust the tonal values of the image.  Figure~\ref{fig:photofinishing} shows an example.  Tone mapping serves two purposes.  The first is combined with color manipulation to adjust the image's aesthetic appeal, often by increasing the contrast.  Second, the final output image is usually only 8 to 10 bits per channel (i.e., 256 or 1024 tonal values) while the raw-RGB sensor represents a pixel's digital value using 10--14 bits (i.e., 1024 up to 16384 tonal values). As a result, it is necessary to compress the tonal values from the wider tonal range to a tighter range via tone mapping.  This adjustment is reminiscent of the human  eye's adaptation to scene brightness~\citep{Land:Retinex}. Figure~\ref{fig:tone-map} shows a typical 1D LUT used for tone mapping.

\paragraph*{Noise reduction:}~~Noise reduction algorithms are a key step to improving the visual quality of the image. A delicate balance must be struck in removing image noise while avoiding the suppression of fine-detail image content.  Too aggressive denoising and the image may have a blurred appearance.  Too little image denoising may result in visual noise being dominant and distracting in the final image. Given the importance of denoising, there is a large body of literature on this problem, which we will discuss in detail in Section~\ref{sec:denoising}. Denoising algorithms often consider multiple factors, including the captured image's ISO gain level and exposure settings.  While we show noise reduction applied after the color space transform and photo-finishing, some ISPs apply noise reduction before photo-finishing, or both before and after. Indeed, ISPs often provide denoising algorithms that can be tuned by camera makers to taste.

\paragraph*{Output color space conversion:}~~At this stage in the camera pipeline the image's RGB values are in a wide-gamut ProPhoto color space.  However, modern display devices can only produce a rather limited range of colors.  As a result, the image is converted to a display-referred (or output-referred) color space intended for consumer display devices with a narrow color gamut.  The most common color space is the standard RGB (sRGB)\footnote{which is also the standard for HDTV.}.  Other color spaces, such as AdobeRGB and Display-P3, are sometimes used. The output color space conversion includes a tone-mapping operator as part of its color space definition.  This final tone-mapping operator is referred to as a ``gamma'' encoding.  The name comes from the Greek letter used in the formula to model the nonlinear tone curve.  The purpose of the gamma encoding is to code the digital values of the image into a perceptually uniform domain~\citep{Poynton:2012:Book-DigitalVideo}.  The gamma values used for sRGB and Display-P3 closely follow Stevens's power law coefficients for perceived brightness~\citep{Stevens:Science}.

\paragraph*{Image resizing:}~~The image can be resized based on the user preferences or target output device (e.g., if the camera is used in a preview mode with a viewfinder).  Image resizing is not limited to image downsizing, but can be employed to upsample a cropped region in the captured image to a larger size to provide a ``digital zoom.''  More details of this operation appear in Section \ref{sec:upscaling}.

\paragraph*{JPEG compression and metadata:}~~The image is finally compressed, typically with the JPEG compression standard, and saved. Additional information, such as capture time, GPS location and exposure setting, can be saved with the image as metadata.

\subsection{Modern multi-frame (burst) pipeline}
\label{sec:burst-pipeline}
The advancement of smartphone displays, and networking capabilities in 2010, together with the emergence of image sharing services, like Instagram and Pinterest, resulted in users taking many more photographs with their smartphones and sharing them more broadly. Seeing their photographs mixed alongside professionally produced content for the first time, users began demanding higher quality from their smartphone cameras. This increased demand for higher quality spurred innovation in the industry. Smartphone manufacturers began utilizing the additional computing power of the smartphone by incorporating computational photography techniques in an effort to overcome the limitations enforced by the smartphone's physical form and to bridge the quality gap relative to dedicated camera devices, like DSLRs.

\begin{figure}[h]
\centering
\includegraphics[width=5in]{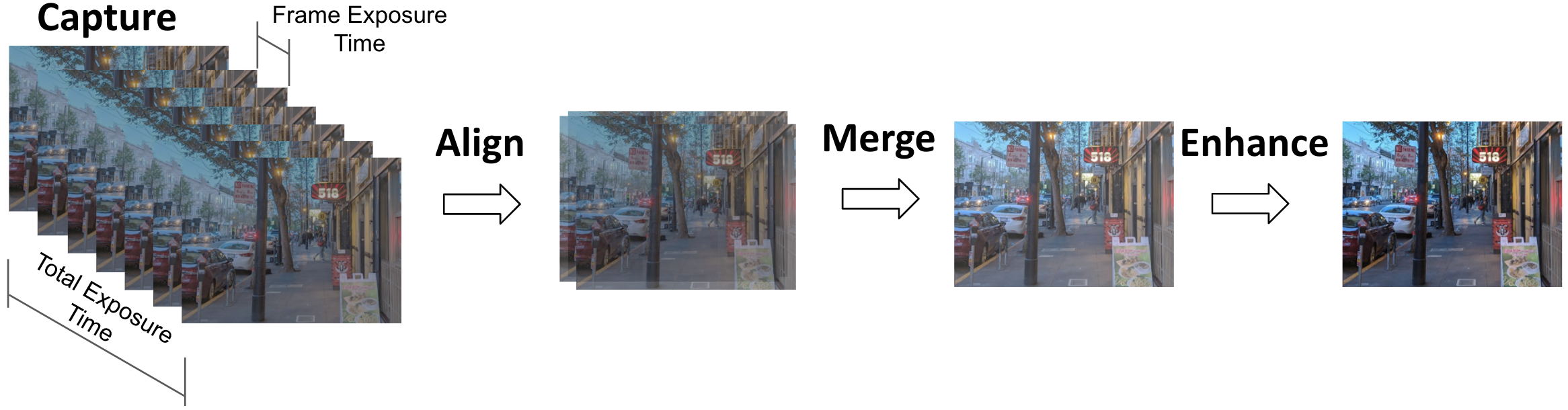}
\caption{Burst photography used in most mobile imaging pipelines consists of four major steps: \textbf{Capture:} A burst of frames is captured based on an exposure schedule defining the total number of frames to capture as well as the exposure time for each individual frame. This defines the total exposure time for the burst. \textbf{Align:} Frames in the burst are spatially aligned. \textbf{Merge:} The aligned frames are merged into a single output frame. \textbf{Enhancement:} Encapsulates all post-processing steps after the merge step, including color manipulation, tone mapping and noise reduction.} 
\label{fig:burst}
\end{figure}

Over the course of the past decade, the modern smartphone camera pipeline has evolved around the concept of capturing and merging multiple frames (known as {\em burst processing}) to generate images of greater quality than possible through the capture of a single image alone. This approach has a comparatively long history in the computational photography research domain, where multi-frame processes have been proposed for denoising \citep{Mildenhall:2018:BDK, Godard:2018:DBD}, joint demosaicing and denoising  \citep{Tan:2017:JDD, Gharbi:2016:DJD}, joint demosaicing and super-resolution \citep{Farsiu:2006:MDS}, and high-dynamic-range imaging \citep{debevec1997, mertens2007}.
Recent years have seen the incorporation of multi-frame techniques like these into the default photography mode of smartphone cameras aimed at synthesizing higher-resolution sensors with larger pixels and higher bit depth.

To implement burst processing, smartphones can incorporate a strategy of continuous capture, where on launching the camera, frames are continuously captured and stored into a ring buffer\footnote{~A \hyperlink{https://en.wikipedia.org/wiki/Circular_buffer}{ring buffer} stores a sequence of frames in order and has the property that when it is full and a subsequent new frame is captured, this over-writes the oldest frame in the buffer, keeping a constant number of frames in memory.}. In this mode, known as \emph{zero shutter lag} (ZSL), on a shutter press, the buffer of frames is passed to the camera processing pipeline for merging. The merge process selects a single frame from the burst close to the shutter press as a ``base'' frame, and aligns and merges information from surrounding frames to improve image quality. Two critical factors in this process are the correct exposure of captured frames to ensure adequate light capture with minimal motion blur, and accurate motion alignment between frames. In this way, the processing pipeline aims to loosely emulate our own ability to annul image motion by adaptation of our spatiotemporal receptive fields \citep{Burr:1986}. The general structure shown in Figure~\ref{fig:burst} illustrates the modern burst processing pipeline used in many mobile cameras, consisting of exposure control, frame alignment, merging, and post-merge image enhancement components. We explore these components next. 

\subsubsection{Exposure Control}
Collecting sufficient light through accurate exposure is critical in controlling image noise, blur, and dynamic range for high-quality photography~\citep{Hasinoff:2010:CVPR}. The signal-to-noise ratio varies proportionally to the exposure time, so for shadow regions or dimly lit scenes, it is important to set the exposure high enough to suppress noise and capture sufficient image detail. However, setting the exposure too high can cause the captured light to exceed the image sensor’s pixel well capacity, resulting in saturated (or blown-out) image details.

Exposure can be adjusted through the combination of three camera settings: the aperture, ISO (sensor gain / effective sensor sensitivity), and the shutter speed (exposure time). However, given that most smartphones have a fixed aperture, exposure control is typically limited to adjustment of the sensor gain and exposure time. Some smartphone cameras do provide manual control over the exposure, but most target the non-expert user and instead control the exposure automatically.

Approaches to estimating the correct exposure time for a single captured image often use measurements such as the distance between the maximum and minimum luminance as well as the average luminance in the scene~\citep{sampat:auto_focus}. In a smartphone’s burst processing pipeline, however, exposure control is more complex since the captured image is generated from merging multiple frames. For burst processing, the camera must define a schedule of exposure times for a burst of captured frames to achieve an overall total exposure time for the merged output frame (see Figure~\ref{fig:burst}). For example, a total exposure time of 300ms could be achieved through a schedule of five frames, each with an exposure time of 60ms. Similarly, in a burst processing pipeline implementing HDR through bracketing, the exposure schedule might define a schedule of short, medium and long exposures. Exposure control for burst processing therefore needs to take into consideration not only the available light in the scene but also the merge processing and how it impacts the overall exposure.

Another factor that greatly impacts exposure control is camera shake (e.g., due to being handheld), which can introduce motion blur. To enable longer exposure times, modern smartphone cameras incorporate an optical image stabilizer (OIS), which actively counteracts camera shake. However, this often does not completely remove the motion and does not help in the case of (local) subject motion, which is also a source of motion blur. Adapting the exposure schedule in accordance with motion observed in the scene is a common approach used to reduce the impact of motion blur. In Section~\ref{sec:low-light-imaging} we further examine exposure control and more severe limitations in the case of low-light photography.

\subsubsection{Alignment}

Generating high-quality, artifact-free, images through burst processing relies on robust and accurate spatial alignment of frames in the captured burst. This alignment process must account for not only  global camera motion (residual motion not compensated for by the OIS) but also local motion in the scene. There is a long history of frame alignment techniques in the research literature, from early variational methods that solve the global alignment problem using assumptions of brightness constancy and spatial smoothness \citep{horn1981determining}, to multi-scale approaches solving for both global and local motion \citep{Bruhn:2005}. 

Given the omnipresent convenience of the smartphone, photography has been made possible in the most extreme of conditions. As a result, accurate alignment can be challenging, particularly in under-exposed or low-light scenes, where noise can dominate the signal. Similarly, over-exposed scenes can introduce clipping or motion blur, making alignment difficult or impossible due to the loss of image detail. Even with optimal exposure, complex non-rigid motion, lighting changes, and occlusions can make alignment challenging.

Although state-of-the-art multi-scale deep learning methods can achieve accurate frame alignment in challenging conditions \citep{Sun:2018}, they are currently beyond the computational capabilities of many smartphones. As a result, burst processing on a smartphone is greatly limited by the accuracy of the alignment process. The exposure schedule must be defined so as to facilitate accurate alignment, and the merge method must in turn be designed to be robust to misalignments to avoid jarring artifacts in the merged output (also known as fusion artifacts). Common merge artifacts due to misalignments include \emph{ghosting} and \emph{zipper} artifacts, often observed along the edges of moving objects in a captured scene. 

\subsubsection{Merge}
Once accurately aligned, the smartphone’s burst processing pipeline must reduce the multiple captured frames to a single output frame. In static scenes and in the absence of camera shake, a simple averaging of frames of the same exposure will reduce noise proportionally to the square root of the total number of merged frames. However, few scenarios arise in real-world photography where such a simple merging strategy could be effectively applied. Also, such a simple merge strategy under-utilizes the burst processing approach, which, as previously mentioned, can also facilitate increasing dynamic range and resolution. In this section, we describe a merge method aimed at reducing noise and increasing dynamic range called HDR+~\citep{Hasinoff:2016:BPH}, but later in Section~\ref{sec:super-resolution} we describe a generalization of the method aimed at increasing resolution as well.

HDR+ was one of the earliest burst processing approaches that saw mass commercial distribution, featuring in the native camera app of Google’s Nexus and Pixel smartphones. Aimed at reducing noise and increasing dynamic range, the HDR+ method employs a robust multi-frame merge process operating on 2--8 frames, achieving interactive end-to-end processing rates. To reduce the impact of motion blur and to avoid pixel saturation, HDR+ defines an exposure schedule to deliberately under-expose the captured frames in a ZSL buffer. Bypassing the smartphone’s standard (single-frame) ISP, the merge pipeline operates on the raw Bayer frames directly from the camera’s sensor, enabling the merge process to benefit from higher bit-depth accuracy and simplifying the modeling of noise in the pipeline.

Given a reference frame close (in time) to the shutter press, the HDR+ pipeline successively aligns and merges alternative frames in the burst, pair-wise. The merging of frame content operates on tiles and is implemented in the frequency domain. For each reference and alternate tile pair, a new tile is linearly interpolated between them (per frequency) and averaged with the reference tile to generate the merged tile output. Given that the merge strategy is applied per frequency, the merging achieved per tile can be partial. The interpolation weight is defined as a function of the measured difference between the aligned tile pairs and the expected (i.e., modeled) noise. For very large measured differences (e.g., possibly due to misalignment), the interpolated output tends towards the reference tile, whereas for differences much less than the expected noise, the interpolated output tends towards the alternate tile. By adapting in this way, the merging process provides some robustness to misalignment, and degrades gracefully to outputting the reference frame only in cases where misalignment occurs across the entire burst.

The final output of the merge process is a Bayer frame with higher bit depth and overall SNR, which is then passed to a post-merge enhancement stage, including demosaicing, color correction, and photo-finishing. Of particular importance among these post-merge processes is spatial denoising. As a consequence of the tile-wise, and partial merging of frames, the resulting merged frame can have spatially varying noise strength which must be adequately handled by the post-merge denoising process.

\subsection{Photo-finishing: Denoising, tone mapping, and sharpening}

While significantly improved quality results from merging multiple frames, it is nevertheless the case that, just like the single-frame pipeline, a number of additional {\em photo-finishing} operations are still required to give the final picture the desired aesthetic qualities of a pleasing photograph. These steps (denoising, tone mapping, and sharpening) are some of the core technologies of traditional image processing, and have been studied extensively over several decades preceding the advent of what we now call computational photography. But with the heightened demands of new sensors on smaller devices, the development of these (single-image) enhancement techniques too has been accelerated in recent years. Indeed, the multi-frame approach may require aspects of photo-finishing to be tailored for the multi-frame merged output. As discussed in Section~\ref{sec:hardware_and_limitations}, denoising is an operation of great importance in the pipeline (whether single or multi-frame), and hencewe begin by describing this operation in some detail.

\subsubsection{Denoising} \label{sec:denoising}

 As should be clear from the preceding material, filtering an image is a fundamental operation throughout the computational photography pipeline. Within this class the most widely used canonical filtering operation is one that \emph{smooths} an image or, more specifically, removes or attenuates the effect of noise.  

The basic design and analysis of image denoising operations have informed a very large part of the image processing literature \citep{milanfar2013tour,lebrun2012secrets,hiro_robust,takeda07}, and the resulting techniques have often quickly spread or been generalized to address a wider range of restoration and reconstruction problems in imaging. 

\begin{figure}[h]
\centering
\includegraphics[width=5in]{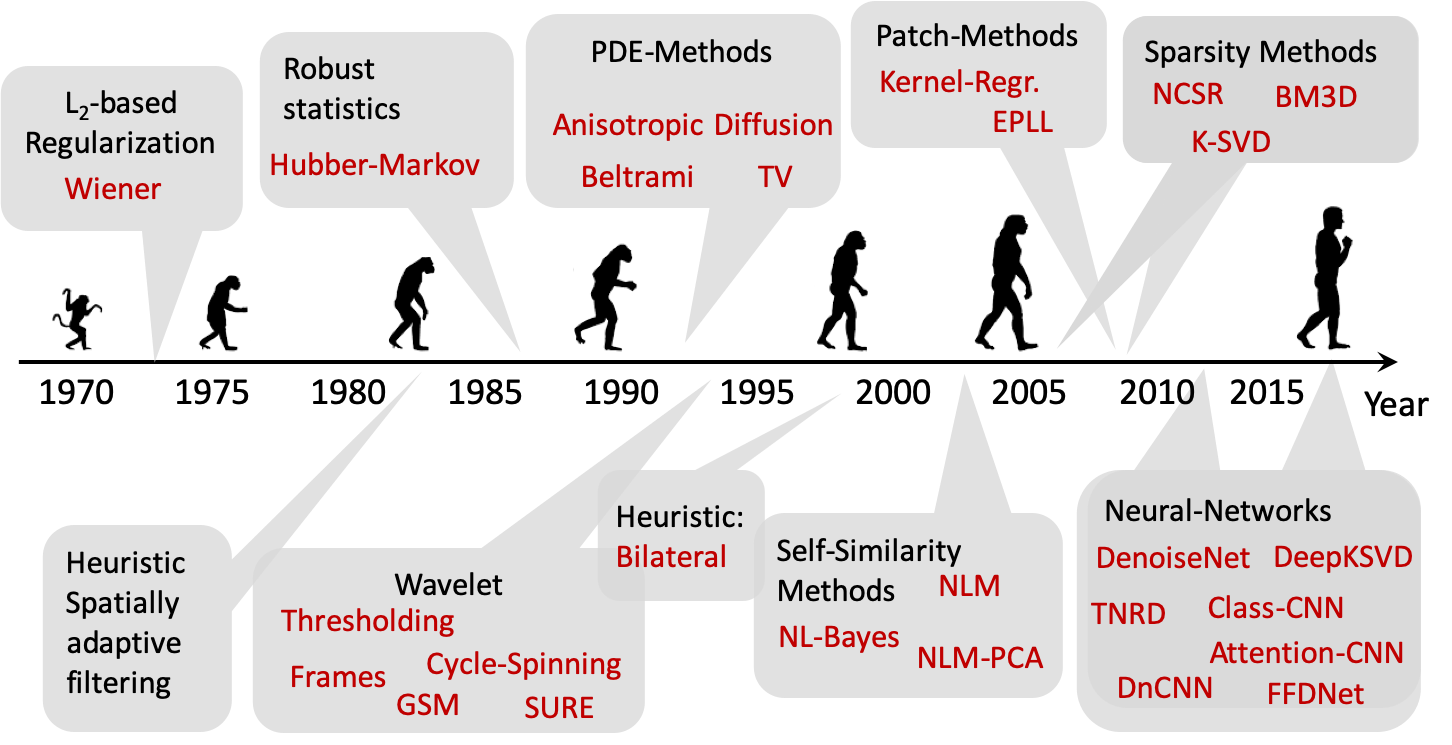}
\caption{The historical timeline of progress in image denoising.} 
\label{fig:progress}
\end{figure}

Over the last five decades, many approaches have been tried, starting with the simplest averaging filters and moving to ones that adapted somewhat better (but still rather empirically) to the content of the given image. 
With shrinking device sizes, and the rise in the number of pixels per unit area of the sensor, modern mobile cameras have become increasingly prone to noise. The manufacturers of these devices, therefore, depend heavily on image denoising algorithms to reduce the spurious effects of noise. A summary timeline of denoising methods is illustrated in Figure \ref{fig:progress}. 

Only relatively recently in the last decade or so (and concomitant with the broad proliferation of mobile devices) has a great leap forward in denoising performance been realized~\citep{chatterjee2010denoising}. What ignited this recent progress were patch-based methods~\citep{efros1999texture,buades_nonlocal}. This generation of algorithms exploits both local and non-local redundancies or “self-similarities” in the image. This now-commonplace notion is to measure and make use of affinities (or similarities) between a given pixel or patch of interest, and other pixels or patches in the given image. These similarities are then used in a filtering (e.g., data-dependent weighted-averaging) context to give higher weights to contributions from more similar data values, and to properly discount data points that are less similar.

Early on, the bilateral filter \citep{tomasi1998bilateral} was developed with very much this idea in mind, as were its spiritually close predecessors like the Susan filter \citep{smith1997susan}. More recent extensions of these ideas include \citep{bm3d,buades_nonlocal,hiro_robust, Zoran_Weiss_EPLL_2011}, and other generalizations described in ~\citep{milanfar2013tour}.

The general construction of many denoising filters begins by specifying a (symmetric positive semi-definite) kernel $k_{ij}(\by) = K(y_i,y_j) \geq 0$, from which the coefficients of the adaptive filter are constructed. Here $\by$ denotes the noisy image, and $y_i$ and $y_j$ denote pixels at locations $i$ and $j$ respectively\footnote{~In practice, it is commonplace to compute the kernel not on the original noisy $\by$, but on a ``pre-filtered'' version of it, processed with some basic smoothing, with the intent to weaken the dependence of the filter coefficients on noise.}. 

Specifically,
\[ w_{ij} = \frac{k_{ij}}{\sum_{i=1}^n k_{ij}}. \]
Each pixel of the denoised image $\widehat{\by}$ is then given by
\[\widehat{y}_j = \sum_{i=1}^n w_{ij} \; y_i,\]
where the coefficients $\left[a_{1j},\;\cdots,\; a_{nj} \right]$ describe the relative contribution of the input (noisy) pixels to the output pixels, with the constraint that they sum to one:
\[\sum_{i=1}^n a_{ij}=1.\]

Let's concretely highlight a few such kernels which lead to popular denoising/smoothing filters. These filters are commonly used in the computational photography, imaging, computer vision, and graphics literature for many purposes.

\subsubsection{Classical Gaussian filters }
Measuring (only) the spatial distance between two pixels located at (2D) spatial positions $x_i$ and $x_j$, the classical Gaussian kernel is
\[
k_{ij} = \exp\left(\frac{-\|x_i-x_j\|^2}{h^2}\right).
\]
Such kernels lead to the classical and well-worn Gaussian filters that apply the same weights regardless of the underlying pixels.

\subsubsection{The bilateral filter} This filter takes into account both the spatial \textit{and} data-wise distances between two samples, in separable fashion, per \cite{tomasi} and \cite{elad1}:
\[ 
 \label{eq:bilat}
k_{ij} = \exp \left(\frac{-\|x_i-x_j\|^2}{h_x^2}\right)\,\exp\left(\frac{-(y_i-y_j)^2}{h_y^2}\right) = \exp\left\{\frac{-\|x_i-x_j\|^2}{h_x^2} + \frac{-(y_i-y_j)^2}{h_y^2} \right\}.
\] 

As can be observed in the exponent on the right-hand side, the similarity metric here is a weighted Euclidean distance between the vectors $(x_i,y_i)$ and $(x_j,y_j)$. This approach has several advantages. Namely, while the kernel is easy to construct, and computationally simple to calculate, it yields useful local adaptivity to the pixels.

\subsubsection{Non-local means}

The non-local means algorithm, originally proposed in \cite{buades_nonlocal}, is a generalization of the bilateral filter in which the data-dependent distance term (\ref{eq:bilat}) is measured block-wise instead of point-wise:
\[ \label{eq:nlm}
k_{ij} = \exp\left(\frac{-\|x_i-x_j\|^2}{h_x^2}\right)\,\exp\left(\frac{-\|\by_i-\by_j\|^2}{h_y^2}\right),
\]
where $\by_i$ and $\by_j$ refer now to patches ({\em subsets} of pixels) in $\by$.

\subsubsection{Locally adaptive regression kernel}
The key idea behind this kernel is to robustly measure the local structure of data by making use of an estimate of the local {\em geodesic} distance between nearby samples, \cite{takeda07}:
\[ \label{LARK}
k_{ij} = \exp \left\{ -(x_i - x_j)^T
\mathbf{Q}_{ij} (x_i - x_j) \right\},
\]
where $\mathbf{Q}_{ij} = \mathbf{Q}(y_i,y_j)$ is the covariance matrix of the gradient of sample values estimated from the given pixels, yielding an approximation of local geodesic distance in the exponent. The dependence of $\mathbf{Q}_{ij}$ on the given data means that the denoiser is highly nonlinear and shift varying. This kernel is closely related, but somewhat more general than the Beltrami kernel of \cite{kimmel1} and the coherence-enhancing diffusion approach of \cite{weickert}.

\vspace{.1in}

More recently, methods based on deep convolutional neural networks (CNNs) have become dominant in terms of the quality of the overall results with respect to well-established quantitative metrics ~\citep{harmeling,meinhardt17learning}. Supervised deep-learning based methods are currently the state of the art---for example~\citep{
harmeling,Chen_TNRD_2015,wang2015deep,Zhang_DnCnn_2017,Mao_RED_2016,Zhang_FFDNet_2018,remez2018class,Tai_MemNet2017,Liu_MwCnn2018,Zhang_RDN_2018,Liu_NLRN_2018}. However, these CNN-based approaches are yet to become practical (especially in mobile devices) due not only to heavy computation and memory demands, but their tendency to sometimes also produce artifacts that are unrealistic with respect to more qualitative perceptual measures.  

Finally, it is worth noting that as denoising methods evolve from traditional signal/image-processing approaches to deep neural networks, there has been an increasing need for training sets comprised of images that accurately represent noise found on small sensors used in camera phones.  Towards this goal, the recent DND~\citep{plotz2017benchmarking} and SSID datasets~\citep{abdelhamedCVPR2018SSID} provide images captured directly from such devices for use in DNN training.  Both works have shown that training using real images vs. those synthesized from existing noise models provides improved performance.  This hints at the need for better noise models in the literature that are able to capture the true characteristics of small camera sensors. 

\subsubsection{Tone mapping}

As discussed in Section~\ref{sec:single_frame_pipeleine}, tone mapping is applied as part of the photo-finishing routines in both single-frame and multi-frame camera pipelines.  Tone mapping manipulates the intensity levels, or tones, in the image.  Assume that the variable $r$ represents the intensity levels for one of the color channels (R, G, B) that will be enhanced.  Also, assume that the image intensity values have been normalized such that they lie on the interval $[0,1]$.  A tone map can be expressed as follows:
\[ \label{eq:tonemap}
    s = T(r),
\]
where function $T$ produces a new intensity level $s$ for every input level $r$ (i.e., it maps input tones to output tones).  Tone mapping is applied either globally or locally.  A global tone map, often referred to as a  tone curve, is applied to all pixels' intensity values in the image irrespective of the pixel's spatial location.  A global tone map is constrained to satisfy the following conditions:\\
(1) $T(r)$ is single-valued and monotonically increasing; and \\
(2) $0 \leq T(r) \leq 1$.

Because images have discrete intensity values, a global $T(r)$ can be implemented as a 1D LUT.  A global tone map can be applied to each color channel, or a separate tone map can be designed per color channel.   In addition, tone maps can be customized depending on the mode of imaging.  For example, in burst mode for low-light imaging, a tone map can be adjusted to impart a night scene's look and feel.  This can be done using a tone map that maintains strong contrast with dark shadows and strong highlights~\citep{levoy:night_sight_blogpost}.   

\begin{figure}[h]
\centering
\includegraphics[width=5in]{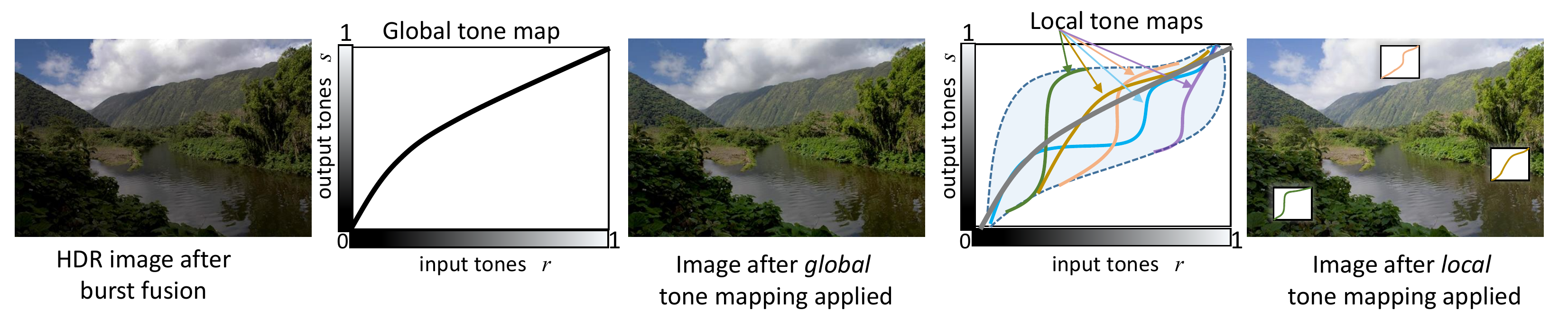}
\caption{An example of a global tone map applied to an HDR image and local tone maps that vary spatially based on the image content.} \label{fig:tone-map}
\end{figure}

Local tone mapping adjusts intensity values in a spatially varying manner. This is inspired by the human visual system's sensitivity to local contrast.   Local tone mapping methods, often referred to as tone operators, examine a spatial neighborhood about a pixel to adjust the intensity value~\citep{Paris:LTM:2011,
Chalmers:Book:2017,
Lee:ICCE:2013,
Reinhard:2002,
Cerda:2018,Wang:TIP:2015}.   As a result, the single-valued and monotonicity constraints are not always enforced as done in global tone mapping.  For example, in the case of burst imaging for HDR, intensity values from the multiple images can be combined in a manner that darkens or lightens regions to enhance the image's visual quality.  Most methods decomposed the input image into a base layer and one or more detail layers.  The detail layers are adjusted based on local contrast, while a global tone map modifies the base layer.  Figure~\ref{fig:tone-map} shows an HDR image that has been processed using both global and local tone mapping.

\subsubsection{Sharpening}
Image sharpness is one of the most important attributes that defines the visual quality of a photograph. Every image processing pipeline has a dedicated component to mitigate the blurriness in the captured image. Although there are several methods that try to quantify the sharpness of a digital image, there is no clear definition that perfectly correlates with the quality perceived by our visual system. This makes it particularly difficult to develop algorithms and properly adjust their parameters in such a way that they produce appealing visual results in the universe of use cases and introduce minimal artifacts.

Image blur can be observed when the camera’s focus is not correctly adjusted, when the objects in the scene appear at different depths, or when the relative motion between the camera and the scene is faster than the shutter speed (motion blur, camera shake). Even when a photograph is perfectly shot, there are unavoidable physical limitations that introduce blur. Light diffraction due to the finite lens aperture, integration of the light in the sensor, and other possible lens aberrations introduce blur, leading to a loss of details. Additionally, other components of the image processing pipeline itself, particularly demosaicing and denoising, introduce blur. 

A powerful yet simple model of blur is to assume that the blurry image is formed by the local average of nearby pixels of a latent unknown sharp image that we would like to estimate. This local average acts as a low-pass filter attenuating the high-frequency image content, introducing blur. This can be formally stated as a convolution operation---that is,
$$
\mathbf{v}[i,j] = \sum_{k,l} h[k,l]~ \mathbf{u}[i-k,j-l],
$$ 
where $\mathbf{v}$ is the blurry image that we want to enhance, $\mathbf{u}$ is the ideal sharp image that we don't have access to, and $h$ models the typically unknown blur filter. 

There are two conceptually different approaches to remove image blur and increase  apparent sharpness. Sharpening algorithms seek to directly boost high- and mid-frequency content (e.g., image details, image edges) without explicitly modeling the blurring process. These methods are sometimes also known as edge enhancement algorithms since they mainly increase edge contrast.  On the other hand, de-convolution methods try to explicitly model the blurring process by estimating a blur kernel $h$ and then trying to invert it. In practice, there are an infinite number of possible combinations of $\mathbf{u}$ and $h$ that can lead to the same image $\mathbf{v}$, which implies that recovering $\mathbf{u}$ from $\mathbf{v}$ is an ill-posed problem. One of the infinite possible solutions is indeed the no-blur explanation: $\mathbf{u}$ = $\mathbf{v}$, and $h$ is the trivial kernel that maintains the image unaltered. This implies that the degradation model is not sufficient to disentangle the blur $h$ and the image $\mathbf{u}$ from the input image $\mathbf{v}$, and more information about $h$ and/or $\mathbf{u}$ (prior) is needed.

\begin{figure}
    \centering
    \includegraphics[width=\linewidth, clip, trim=0 10 0 0]{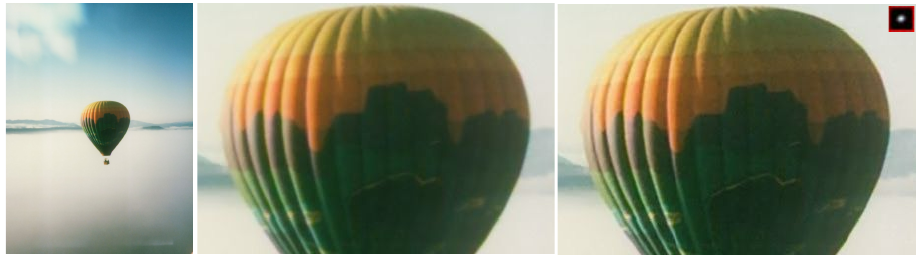}
    \caption{Example of deblurring a mildly blurry image using Polyblur~\cite{delbracio2020polyblur}. The estimated blur kernel is shown on top-right in the right panel. }
    \label{fig:polyblur}
\end{figure}

Most blind deconvolution methods proceed in two steps: a blur kernel is first estimated and then using the estimated kernel a non-blind deconvolution step is applied. These methods generally combine natural image priors (i.e., what characteristics does a natural sharp image have), and assumptions on the blur kernel (e.g., maximum size) to cast the blind deconvolution problem as one of variational optimization~\cite{fergus2006removing,levin2009understanding,el2014comparative}. In the specific case of deblurring slightly blurry images, we can proceed in a more direct way by filtering the image with an estimate of the blur and thus avoid using costly optimization procedures~\cite{hosseini2019convolutional,delbracio2020polyblur}. Figure~\ref{fig:polyblur} shows an example of Polyblur~\cite{delbracio2020polyblur} that efficiently removes blur by estimating the blur and combining multiple applications of the estimated blur to approximate its inverse.

\begin{figure}
\footnotesize
    \centering
    \begin{minipage}[c]{.25\linewidth}
    \centering
    \includegraphics[width=\linewidth]{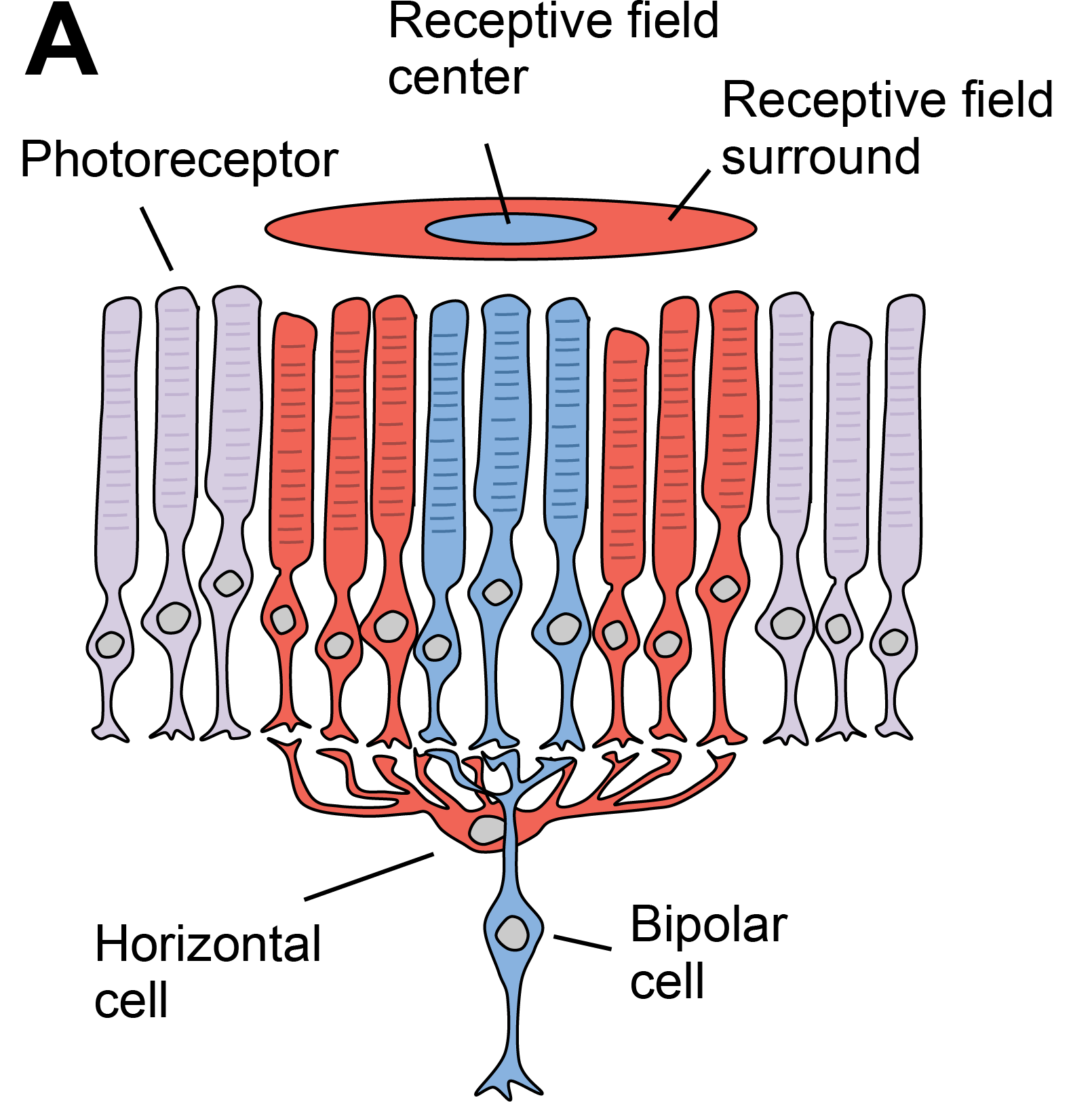} 
    \tinyb{Lateral inhibition in Vertebrates} 
    \tinyb{from~\citep{kramer2015lateral}}

    \end{minipage}
    \begin{minipage}[c]{.2\linewidth}
    \centering
    \includegraphics[width=\linewidth]{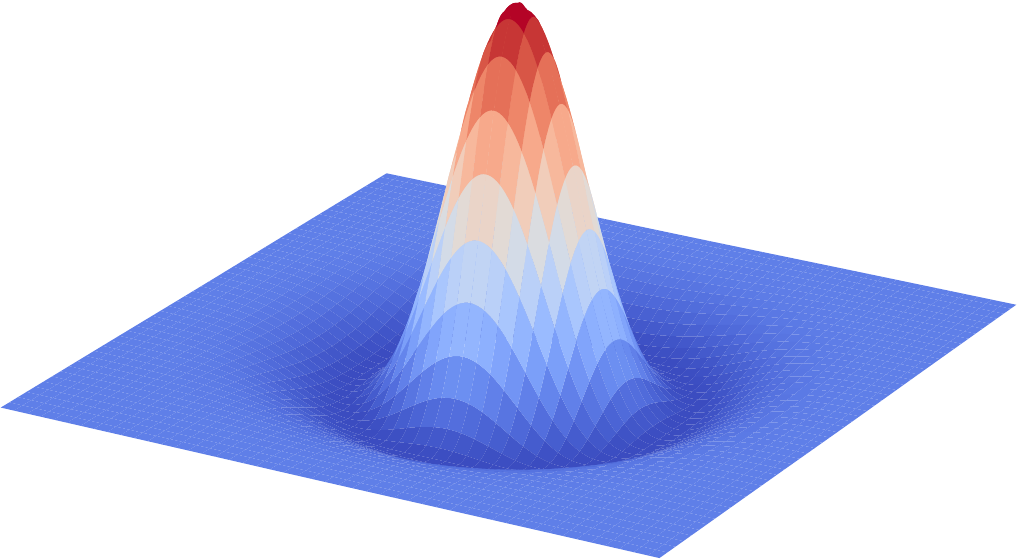} 

    \tinyb{Laplace operator}
    \end{minipage}
    \begin{minipage}[c]{.25\linewidth}
    \centering
    \includegraphics[width=.95\linewidth]{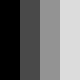} 
    \tinyb{Mach bands}
    \end{minipage}
    \begin{minipage}[c]{.25\linewidth}
    \centering
    \includegraphics[width=.9\linewidth]{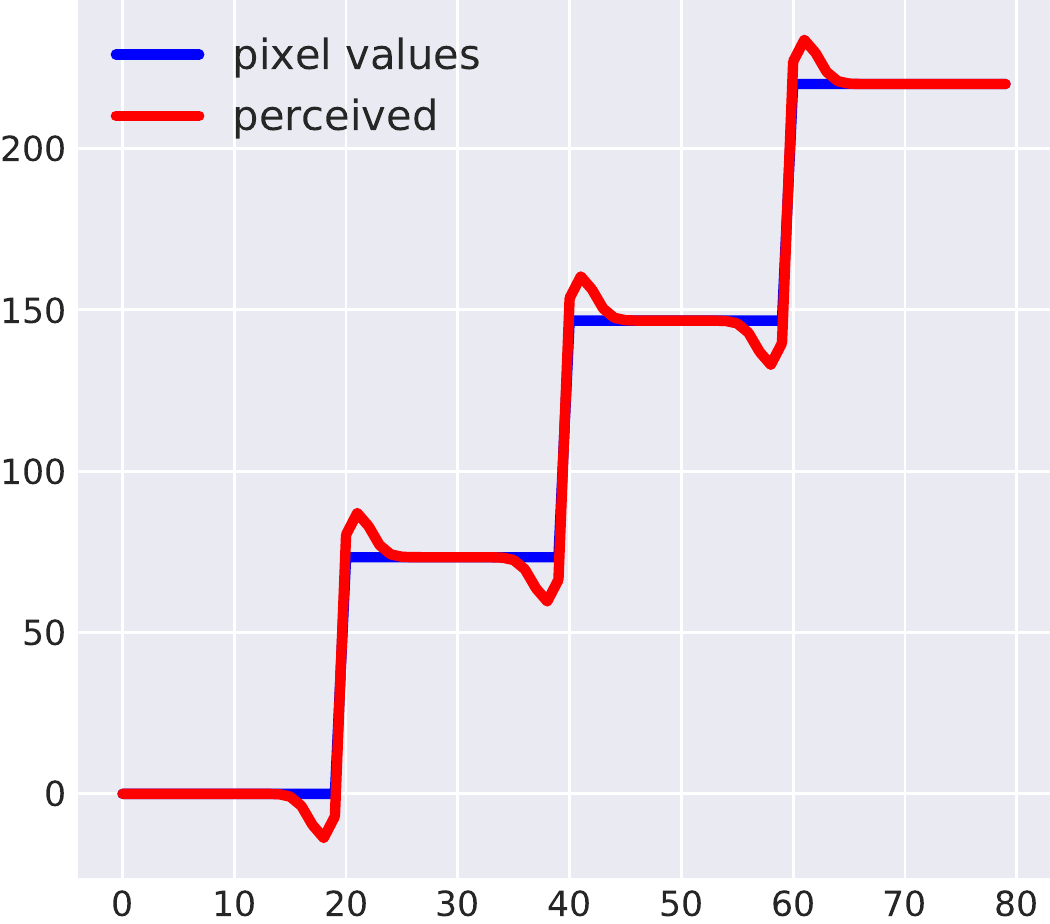} 
    \tinyb{Perceived contrast}
    \end{minipage}
    \vspace{1em}
 
    \caption{\emph{Mach bands:} The human visual system enhances local changes in contrast by exciting and inhibiting regions in a way similar to action of the Laplace operator. Darker (brighter) areas appear even darker (brighter) close to the boundary of two bands.}
    \label{fig:mach}
\end{figure}

One of the most popular and simplest sharpening algorithms is unsharp masking. First, a copy of the given image is further blurred to remove high-frequency image details. This new image is subtracted from the original one to create a residual image that contains only image details. Finally, a fraction of this residual image is added back to the original one, which results in boosting of the high-frequency details. This procedure has its roots in analog photography, where a blurry positive image is combined with the original negative to create a new more contrasted photograph. A typical digital implementation of unsharp masking is by using a Gaussian blur,
$$
\mathbf{u}_\text{unsharp} = \mathbf{u} + \kappa ( \mathbf{u} - G_\sigma \mathbf{u}),
$$
where $G_\sigma$ is the Gaussian blur operator having strength $\sigma$. The parameter $\kappa$ and the amount of blur $\sigma$ should be empirically set.

At a very broad level, the human visual system (HVS) behaves similarly to unsharp masking and the Laplace operator~\citep{ratliff1965mach}.  The center-surround receptive fields present in the eye have both excitatory (center) and inhibitory (surrounding) regions. This leads our visual system to enhance changes in contrast (e.g., edge-detection) by exciting and inhibiting regions in a way similar to action of the Laplace operator. In fact, one manifestation of this phenomenon is the well-known \textit{Mach bands} illusion, where the contrast between edges is exaggerated by the HVS (Figure~\ref{fig:mach}).

There are different variants of this high-frequency boosting principle. \citep{kovasznay1955image} introduced the idea that a mildly blurred image could be deblurred by subtracting a small amount of its Laplacian:
$$
\widehat{\mathbf{u}} = \mathbf{u} - \kappa \Delta \mathbf{u}.
$$
In fact, this method is closely related to unsharp masking, where the residual mask $\mathbf{u} - G_\sigma \mathbf{u}$ is replaced with the negative of the image Laplacian $-\Delta \mathbf{u}$.   

A perhaps not very well-known fact is that the Nobel Prize winner Dennis Gabor studied this process and determined how to set the best amount to subtract~\citep{lindenbaum1994gabor}. In fact, \citep{lindenbaum1994gabor} showed that Laplacian sharpening methods can be interpreted as approximating inverse diffusion processes---for example, by diffusion according to the heat equation, but in reverse time. This connection has led to numerous other sharpening methods in the form of regularized partial differential equations~\citep{osher1990feature, you1996behavioral, buades2006image}.

\section{Compound Features}

A common adage has emerged that describes the age of mobile cameras aptly: ``\emph{The best camera is the one that's with you}.'' The sentiment expressed here declares that there is no longer any need to carry a second camera, if you have a mobile phone in your pocket that has a camera with ``nearly the same functionality.'' Of course, the key caveat is nearly the same. Some of what we take for granted in a large form-factor camera is possible only \emph{because} of the form factor. To approximate those functions we must combine various pieces of technology to emulate the end result. 

Here we briefly describe how we combine the basic techniques described earlier to enable advanced features that not only approximate some functionalities of larger cameras, but also sometimes even exceed them. For instance, the hybrid optical/digital zoom requires state-of-the-art multi-frame merge {\em and} single-frame upscaling technologies. A second example is synthetic bokeh (e.g., synthesizing shallow DoF), which requires both segmentation of the image for depth, and application of different processing to the foreground vs. the background. 

\subsection{Low-light imaging}
\label{sec:low-light-imaging}
When photographing a low-light or night scene, the goal is often not to capture exactly what we see but instead to create a visually pleasing image that also conveys the darkness of the scene. Therefore, unlike human vision, which becomes scotopic in dim light with limited color perception \citep{kelber2017}, smartphone cameras aim to produce photos that are colorful and noise-free. Until relatively recently, high-quality photography in very low-light conditions was  achievable only on standalone cameras like DSLRs, with large enough pixels and adjustable aperture to enable sufficient light capture. As described in Section~\ref{sec:burst-pipeline}, though the exposure time can be increased synthetically by merging multiple frames, there are other factors that inherently limit the maximum achievable exposure time. When operating in a \emph{ZSL} shutter mode (see Section~\ref{sec:burst-pipeline}), the frames acquired by the camera pipeline are also used to drive the viewfinder display. To avoid noticeable judder, the viewfinder must achieve a frame rate of at least 15 frames per second directly, limiting the maximum exposure to 66ms, which is often insufficient for very low-light scenes.

To overcome this, smartphones have adopted a new frame capturing strategy known as positive shutter lag (\emph{PSL}) where frames are captured after the shutter press \citep{levoy:night_sight_blogpost}. By capturing frames after the shutter press, these so-called ``night modes'' can achieve exposure times well beyond the previous 66ms limit. However, to operate robustly in the wild, the camera still needs to automatically adapt to the amount of local and global motion in the scene to avoid the introduction of motion blur. The Night Sight feature of the Pixel smartphone \citep{Liba2019NightSight} solves the motion blur problem through real-time temporal motion metering that runs prior to the shutter press, predicting future scene motion and automatically setting the exposure time and gain (ISO) accordingly, enabling frame exposure times over 300ms in cases where there is little motion.

In addition to the problem of capturing sufficient light, very low-light conditions make tasks such as AWB challenging. \cite{Liba2019NightSight} address this through a learning-based approach to AWB which is trained specifically on  night scenes. \cite{Liba2019NightSight} also introduce a creative tone-mapping solution that draws inspiration from artists' portrayal of night scenes in paintings by keeping shadows close to black and boosting color saturation \citep{levoy:night_sight_blogpost}.

Extending the low-light capabilities of photography even further, some smartphones now offer Astrophotography modes. This has been made possible through more sophisticated motion detection modes that utilize on-device sensors to detect tripod (or non-handheld) mounting, enabling synthetic exposure times of over four minutes \citep{kainz:astro_mode_blogpost}.

\subsection{Super-resolution and hybrid optical/digital zoom}
\label{sec:super-resolution}
Due to the physical limitations on the camera's form factor, one of the principal limitations of a mobile camera is its ability to zoom and focus across a broad range of magnification factors. The thinness of the device prevents the placement of a lens with a broadly variable focal length in front of the sensor. As such, the ability to resolve objects in the mid and far range is inherently limited. 

To address these limitations, two broad classes of approaches have been developed. First, complex optical hardware designs such as the ``periscope lens''\footnote{~\url{https://9to5mac.com/2020/07/22/periscope-lens/}} have enabled larger focal length to be implemented inside typically thin mobile devices. These innovative designs have enabled true optical magnifications as high as $5$ to $10\times$ to be implemented. But the focal length of such telephoto lenses is fixed, and the number of such optical elements is still limited to at most one or two due to the scarcity of space inside the phone, and mechanical limitations. As a result, almost regardless of the optical power of the elements available, the overall zoom pipeline inside all mobile devices has necessarily evolved as a hybrid of optical and digital magnification techniques.

\begin{figure}[h]
\centering
\includegraphics[width=5in]{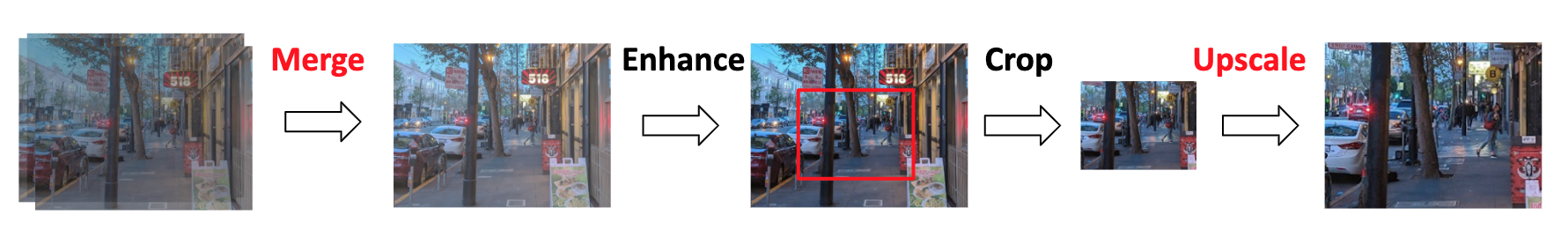}
\caption{The full Super Res Zoom pipeline enhances image resolution in two distinct ways: the merge step, and single-frame upscaling step.} 
\label{fig:SRZoom}
\end{figure}

An example of a modern mobile zoom pipeline is the ``Super Res Zoom'' pipeline described in \citep{SuperresGoogle2019} and used in Google's Pixel devices, illustrated in Figure \ref{fig:SRZoom}. This hybrid optical-digital zoom pipeline first implements a burst processing pipeline that achieves multi-frame super-resolution by aligning, merging, and enhancing a sequence of raw frames with \emph{sub-pixel} accuracy. This process circumvents the need for the typical (first) demosaicing step described earlier in Section \ref{sec:single_frame_pipeleine}.  As a result, the high-frequency (and often aliased) information in the raw frames is used directly in the formation of a full-resolution image at or near the native resolution of the camera.  

This approach implements demosaicing and super-resolution simultaneously, formulating the problem as the reconstruction and interpolation of a continuous signal from a set of possibly sparse samples. Namely, the red, green, and blue pixels measured individually on each frame are reconstructed simultaneously onto a common grid. This technique enables the production of highly detailed images, containing information that would have already been lost in part due to the earlier (and much more naive) interpolation in the demosaicing step. An additional advantage is that it allows us to directly create an image with a desired target magnification / zoom factor.

\begin{figure}[ht]
  \centering
  \includegraphics[width=\linewidth]{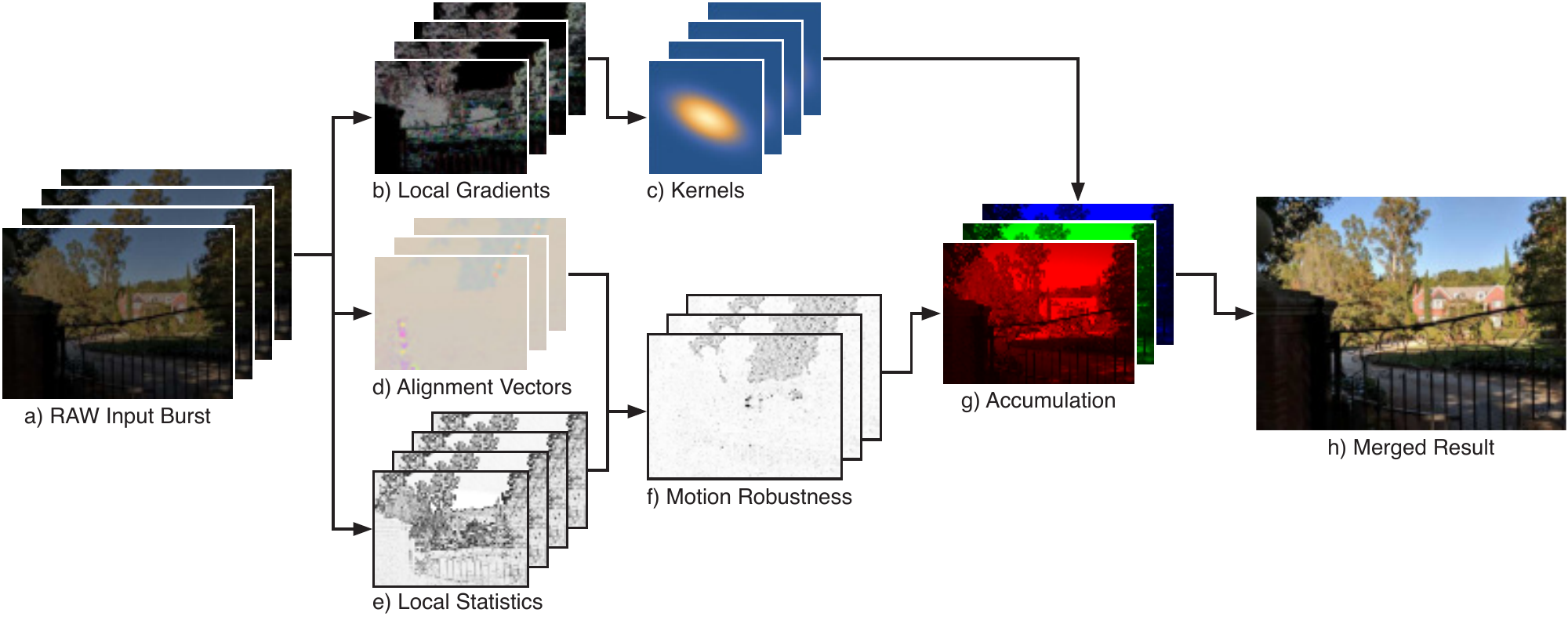}
  \caption{Overview of super-resolution from raw images:
  A captured burst of raw (Bayer CFA) images \textbf{(a)} is the input to our algorithm.
  Every frame is aligned locally \textbf{(d)} to a single frame, called the \emph{base frame}.
  We estimate each frame's contribution at every pixel through kernel regression. \textbf{(g)}.
  The kernel shapes \textbf{(c)} are adjusted based on the estimated local gradients \textbf{(b)} and the sample contributions are weighted based on a robustness model \textbf{(f)}.
  This robustness model computes a per-pixel weight for every frame using the alignment field \textbf{(d)} and local statistics \textbf{(e)} gathered from the neighborhood around each pixel.
  The final merged RGB image \textbf{(h)} is obtained by normalizing the accumulated results per channel.
  We call the steps depicted in \textbf{(b)}--\textbf{(g)} the \emph{merge} step. (Figure from \citep{SuperresGoogle2019})}
  \label{fig:sabre_overview}
\end{figure}

The approach in \citep{SuperresGoogle2019} is visualized in Figure \ref{fig:sabre_overview}. Similar to the standard merge process described earlier, first, a burst of raw (CFA Bayer) images is captured. For every captured frame, it is aligned locally with a single key frame from the burst (called the \emph{base frame}). In the super-resolution application, however, the accuracy of the alignment must meet a higher standard. For instance, due to the color sub-sampling, in order to super-resolve to even the native sensor grid, the accuracy of the registration must be at worst $1/2$ pixel. This is a significantly higher burden both computationally and statistically, making it difficult, if not impossible, to achieve super-resolution in darker scenes, or at high magnification (beyond $2\times$). 
 
With high accuracy alignment in hand, each frame's local contributions are estimated through kernel regression \citep{takeda07} and accumulated across the entire burst, separately per color plane. To achieve local detail, texture, and geometry recovery, the kernel shapes are adjusted based on the estimated signal features and the sample contributions are weighted based on a robustness model (which estimates alignment accuracy). Finally, a per-channel normalization yields the merged RGB image.

Super-resolution is arguably not an alien process to the human visual system. It would appear that the human brain also processes visual stimuli in a way that allows us to discriminate details beyond the physical resolution given by optics and retinal sampling alone. This is commonly known as visual \emph{hyperacuity}, as in \citep{westheimer1975visual}. A possible mechanism of visual super-resolution is the random eye micro-movements known as microsaccades and ocular drifts~\citep{rucci2007miniature,intoy2020finely}. 

\begin{figure}[ht]
  \centering
  \includegraphics[width=.5\linewidth]{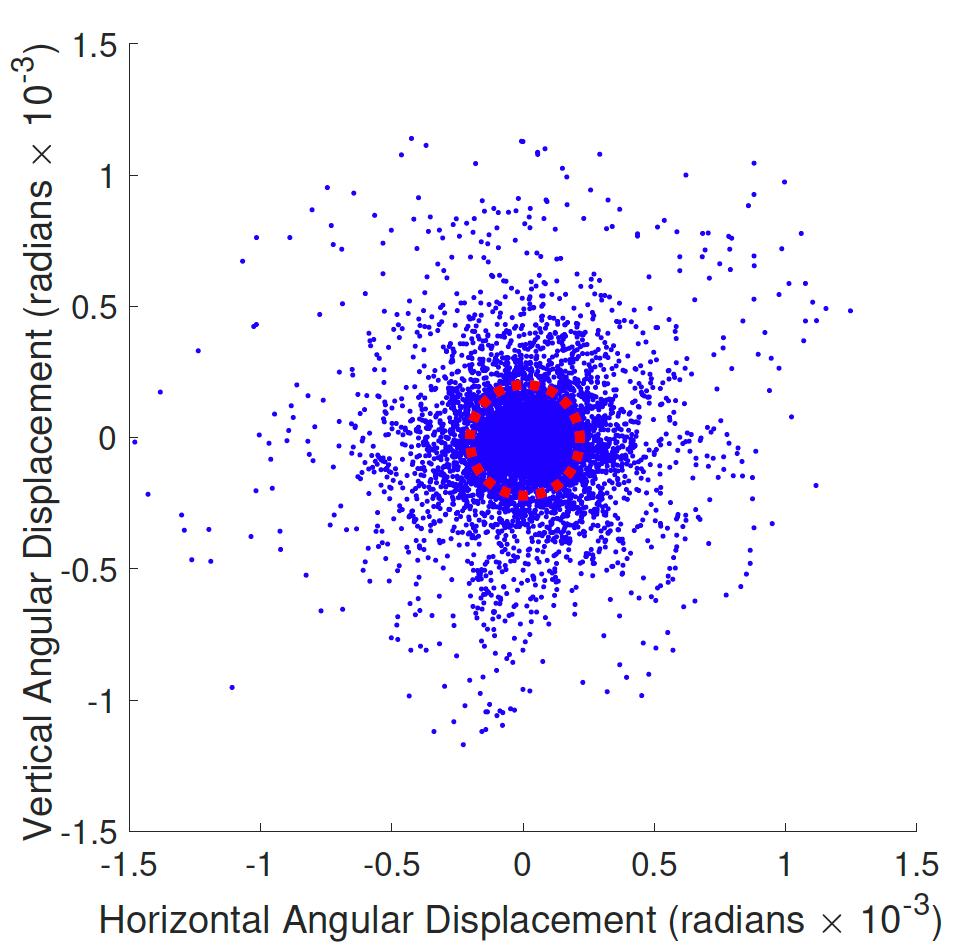}
  \caption{Horizontal and vertical angular displacement (excluding translational displacement) measured from handheld motion across $86$ bursts. Red circle corresponds to one standard deviation, or roughly $0.9$ pixels. (Figure from \citep{SuperresGoogle2019})}
  \label{fig:handheld}
\end{figure}

Interestingly, in the super-resolution work described in \cite{Wronski:2018:SBF}, natural hand tremors play a similar role to eye movements. A natural, involuntary hand tremor is always present when we hold any object. This tremor is comprised of low-amplitude and high-frequency components consisting of a \emph{mechanical-reflex} component, and a second component that causes micro-contractions in the limb muscles~\citep{Riviere:1998:ACP}. In \citep{SuperresGoogle2019}, it was shown that the hand tremor of a user holding a mobile camera is sufficient to provide sub-pixel movements across the images in a burst for the purpose of super-resolution. Experimental measurements of such tremor in captured bursts of images from a mobile device are illustrated in Figure \ref{fig:handheld}.

\begin{figure}[h]
\centering
\includegraphics[width=5in]{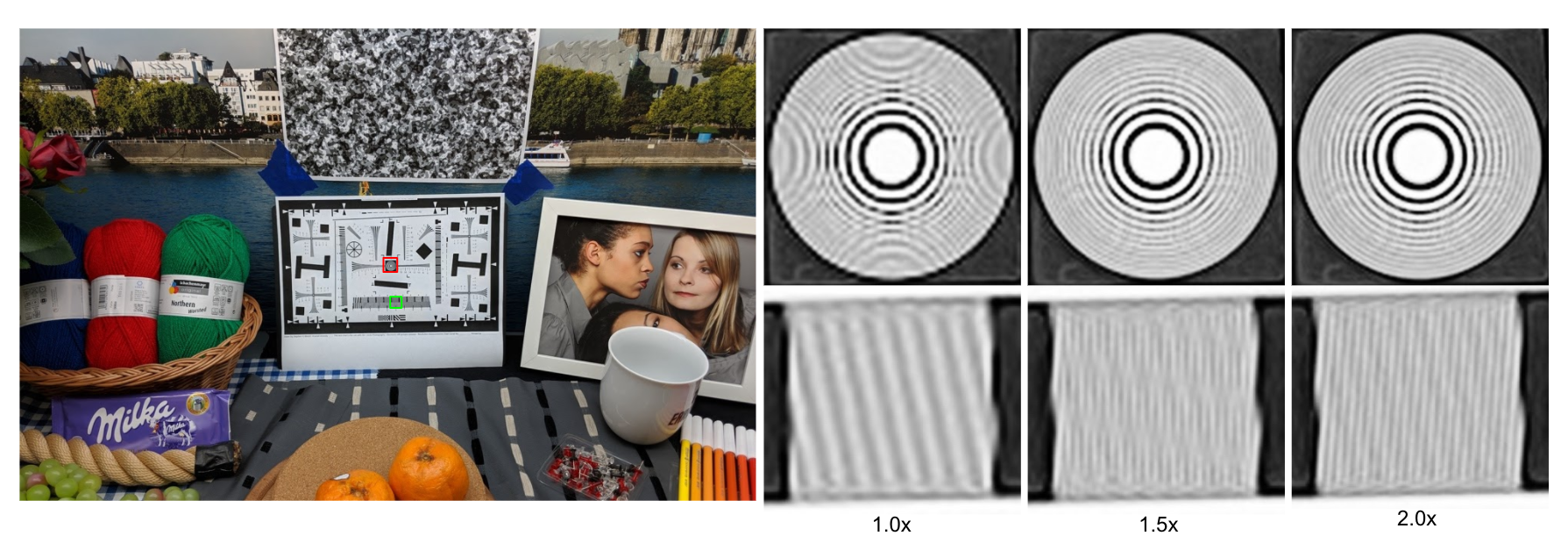}
\caption{Merging a burst onto different target grid resolutions: from left to right, $1 \times$, $1.5 \times$, $2 \times$. The combination of the Super Res Zoom algorithm and the phone's optical system leads to significantly improved results when merged onto a $1.5 \times$ grid; small improvement up to $2 \times$ zoom are also noted. (Figure from \citep{SuperresGoogle2019})} 
\label{fig:SRgrid}
\end{figure}

As illustrated in Figure \ref{fig:SRgrid}, we can see that the merge algorithm alone is able to deliver resolution comparable to roughly a dedicated telephoto lens at a modest magnification factor, no more than $2\times$. Of course the same algorithm can also be applied to the telephoto lens itself, again with typically even more modest gains in resolution.  This suggests that to have a general solution for zoom across a broad range of magnifications, a combination of multi-frame merge and high-quality single-frame crop-and-upscale is required (see Figure \ref{fig:SRZoom}). This upscaling technology is described next.

\subsubsection{Upscaling}\label{sec:upscaling}

Digital zoom, also known as crop-and-scale allows the user to change the field of view (FOV) of the captured photograph through digital post-processing. This operation is essential to allow the photographer to zoom even in cameras that have a dedicated telephoto lens (optical zoom). The operation consists of cropping the image to the desired FOV and then digitally enlarging the cropped region to obtain the desired zoom factor. 

A key challenge of digital zoom lies in performing the upscaling (sometimes also called single-image super-resolution) in a way that preserves image details. Traditional techniques are based on the unrealistic assumption that the digital image is generated by sampling a smooth and regular continuous unknown image. This continuous model allows us to generate arbitrary in-between samples from the observed ones by means of interpolation. There are different interpolation schemes that trade computational cost, quality of the upsampled image (e.g., level of blur introduced), and other possible artifacts. An interpolation scheme is characterized by an interpolation kernel that specifies how the intermediate subpixel sample is computed from the nearby ones.  Bilinear interpolation is simple to compute but generates intermediate samples that result in a final image with blurry appearance.  At the other extreme, the Lanczos interpolation is the one that best approximates the assumed continuous image model but has a higher computational cost since it uses a larger context (large kernel). An advantage of this type of linear interpolation is that intermediate samples can be calculated at arbitrary positions, thus allowing digital zooming of any factor.

\begin{figure}[h]
	\begin{center}
		\subfigure[$2 \times $ upscaling filters]{\includegraphics[width=1\textwidth]{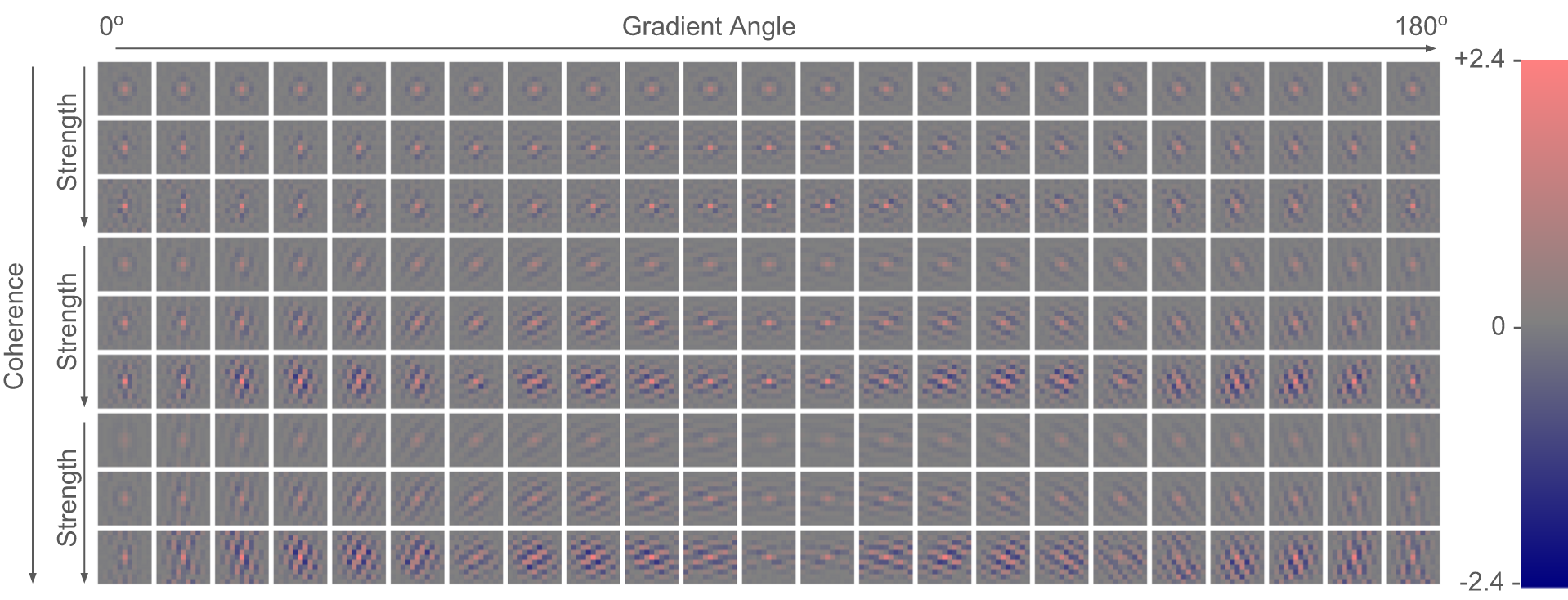}}		
		\subfigure[$3 \times $ upscaling filters]{\includegraphics[width=1\textwidth]{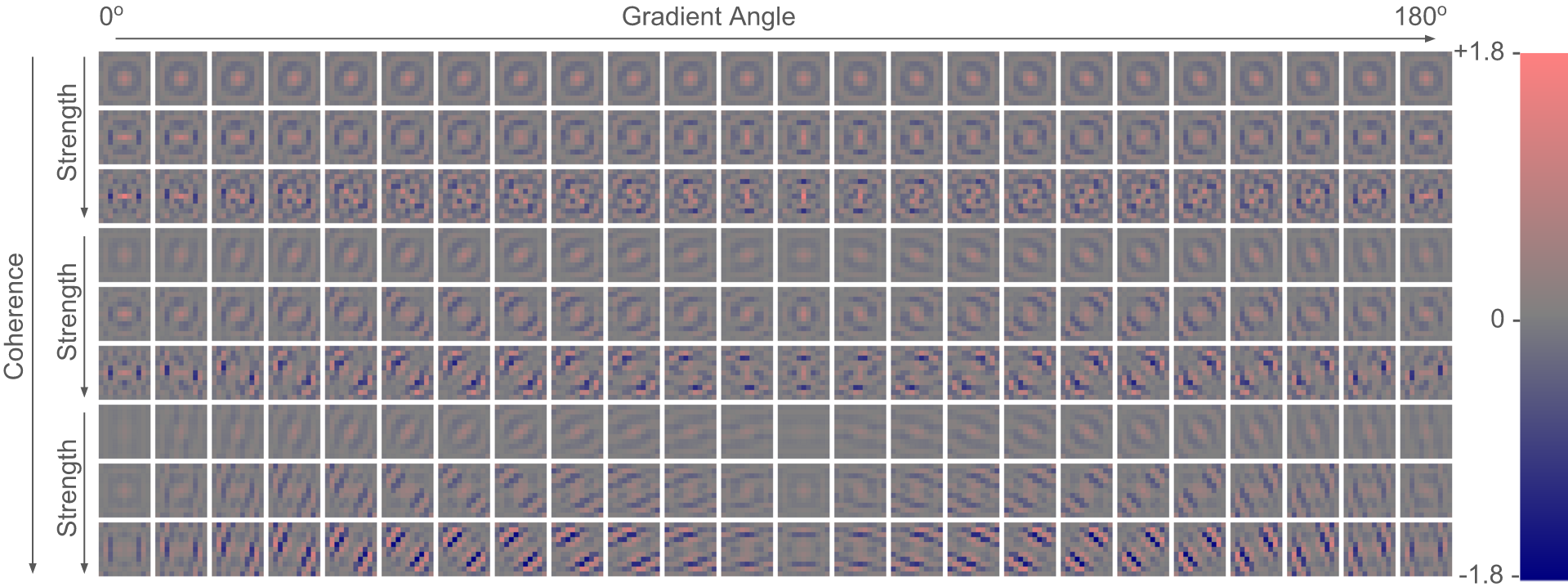}}
	\end{center}
	\caption{Visualization of the learned filter sets for 2$ \times $, and 3$ \times $  upscaling, indexed by angle, strength, and coherence-based hashing of patch gradients. (Figure from \citep{Romano:2016:RAISR})}
	\label{fig:oriented_filters_2x_3x_4x}
\end{figure}

An extension of linear interpolation methods that relies on machine learning techniques is {\it rapid and accurate image super resolution} (RAISR)~\citep{Romano:2016:RAISR}. This method can be seen as a double extension of linear methods. On the one hand, the interpolation kernel is \emph{learned} from a training dataset of pairs of low- and high-resolution images. This is done by finding the best kernel that minimizes the interpolation error for the image pairs in the training dataset. RAISR goes one step further and learns a \emph{collection} of interpolating kernels, each one specialized for a certain local structure encoded by the gradient strength, orientation, and coherence. Examples of such trained filter banks are shown in Figure~\ref{fig:oriented_filters_2x_3x_4x}. Per each subset of filters, the angle varies from left to right; the top, middle, and bottom three rows correspond to low, medium, and high coherence. It is important to note that the filters at different upscaling factors are not trivial transformations of one another. For instance, the 3$\times$ filters are not derived from the 2$ \times$ filters---each set of filters carries novel information from the training data. Given the apparent regularity of these filters, it may also be tempting to imagine that they can be parameterized by known filter types (e.g., Gabor). This is not the case. Specifically, the phase-response of the trained filter is deeply inherited from the training data, and no parametric form has been found that is able to mimic this generally. 
\begin{figure}[h]
	\begin{center}
	\includegraphics[width=.9\textwidth]{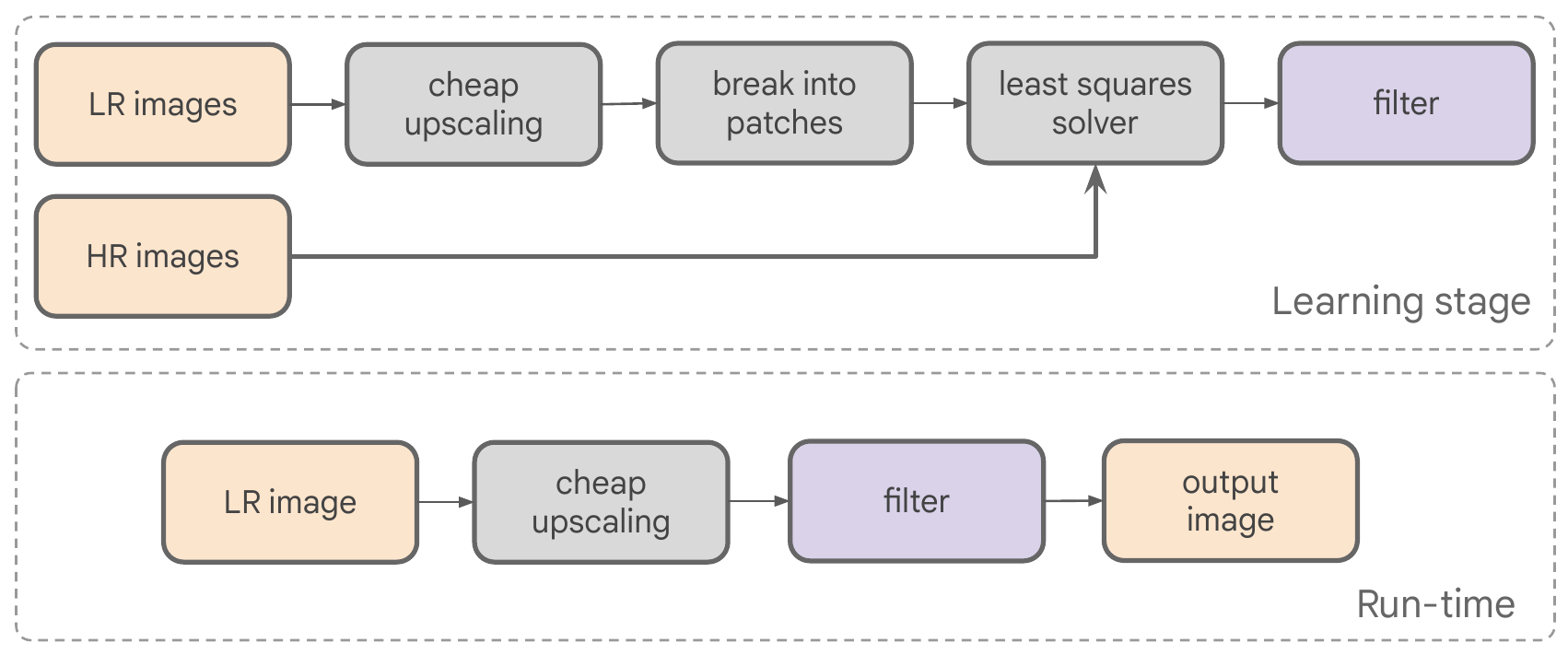}
	\end{center}
	\caption{The learning and application of a filter that maps a class of low-resolution patches to their high-resolution versions. More generally, a set of such filters is learned, indexed by local geometric structures, shown in Figure~\ref{fig:oriented_filters_2x_3x_4x}.}
	\label{fig:globalblock}
\end{figure}

The overall idea behind RAISR and related methods is that well-defined structures, like edges, can be better interpolated if the interpolation kernel makes use of the specific orientation and local structure properties.  During execution, RAISR scans the input image and defines which kernel to use on each pixel and then computes the upscaled image using the selected kernels on a per pixel basis. The overall structure of the RAISR algorithm is shown in Figure \ref{fig:globalblock}. 

RAISR is trained to enlarge an input image by an integer factor (2$\times$--4$\times$), but it does not allow intermediate zooms. In practice, RAISR is combined with a linear interpolation method (bicubic, Lanczos) to generate the zoom factor desired by the user.

With the advancement of deep neural networks in recent years, a wide variety of new image upscaling methods have emerged~\citep{wang2020deep}. Similar to RAISR, deep-learning-based methods propose to learn from image examples  how to map a low-resolution image into a high-resolution one.  These methods generally produce high-quality results but they are not as computationally efficient as shallow interpolation methods, such us RAISR, so their use in mobile phones is not yet widespread. Undoubtedly, deep image upscaling is one of the most active areas of research. Recent progress in academic research, combined with more powerful and dedicated hardware, may produce significant improvements that could be part of the next generation of mobile cameras.
\newpage
\subsection{Synthetic bokeh} \label{sec:synthetic_bokeh}

One of the main characteristics of mobile phone cameras is that the whole image is either in focus or not. The depth of field, defined as the range of depths that are in focus (sharp), is  frequently used by photographers to distinguish the main subject from the background.  As discussed in Section~\ref{sec:hardware_and_limitations}, due to the small and fixed aperture used on smartphone cameras, capturing a shallow DoF image is virtually impossible.

\begin{figure}[h]
\ssmall
    \centering
    \begin{minipage}[c]{.385\linewidth}
    \centering
    \includegraphics[width=\linewidth]{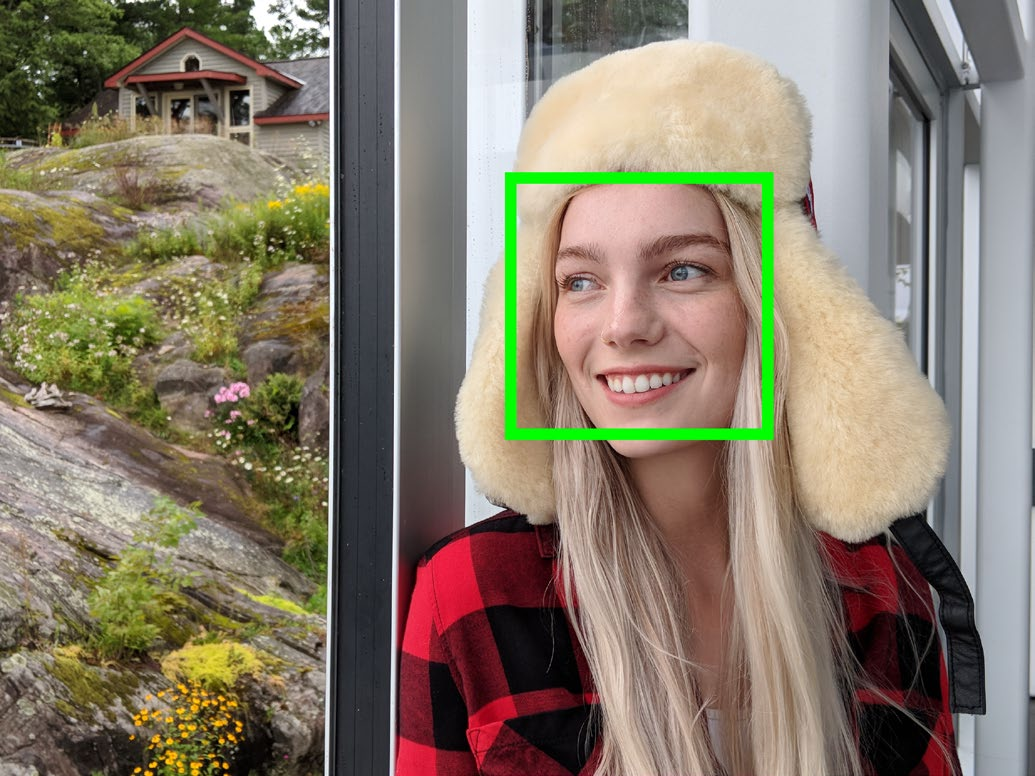}
     Input image with detected face
    \end{minipage}
    \begin{minipage}[c]{.21\linewidth}
    \centering
    \includegraphics[width=.85\linewidth]{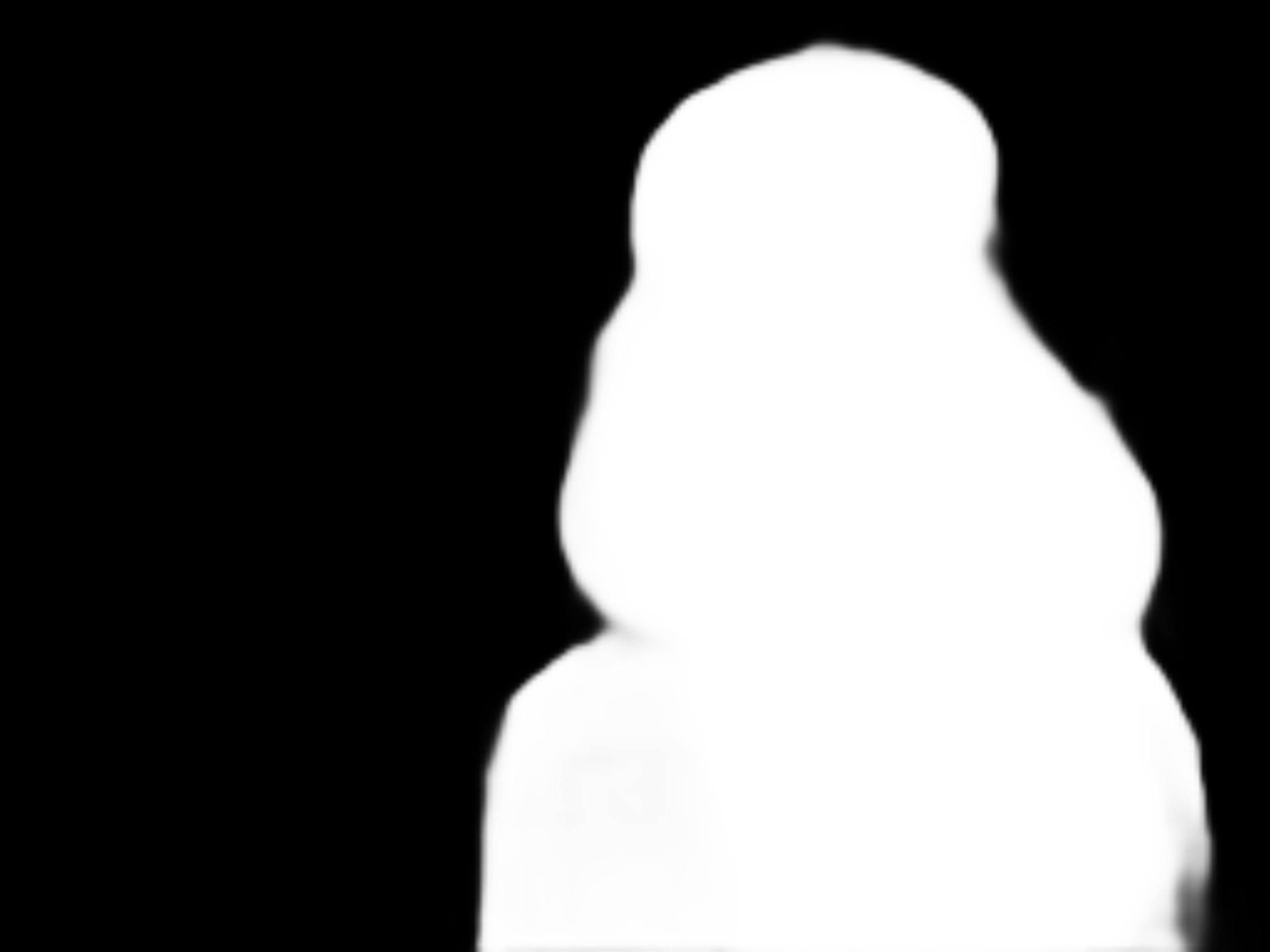}
     Segmentation mask
    \includegraphics[width=.85\linewidth]{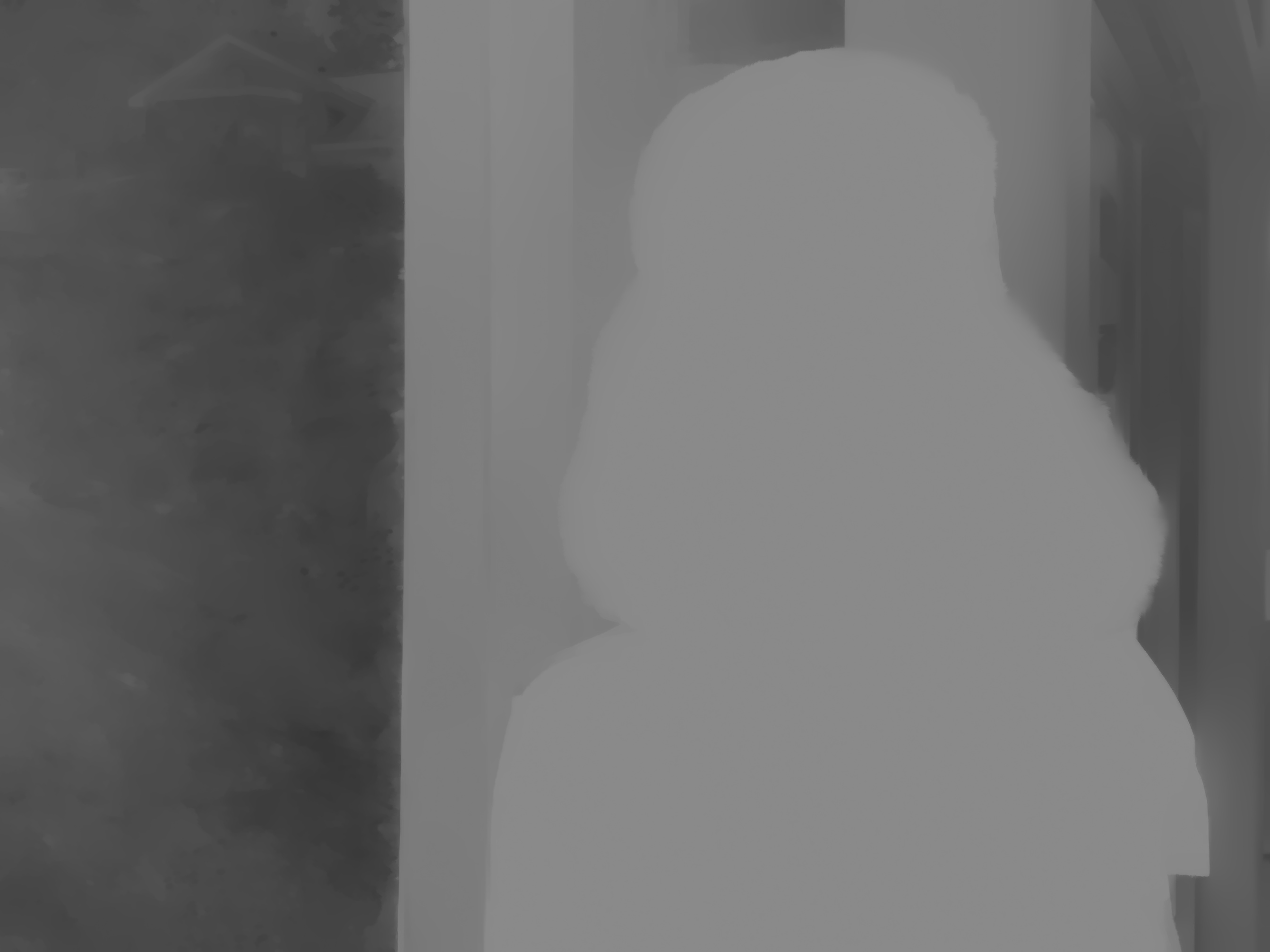}
    Mask + disparity map
    \end{minipage}
    \begin{minipage}[c]{.385\linewidth}
    \centering
    \includegraphics[width=\linewidth]{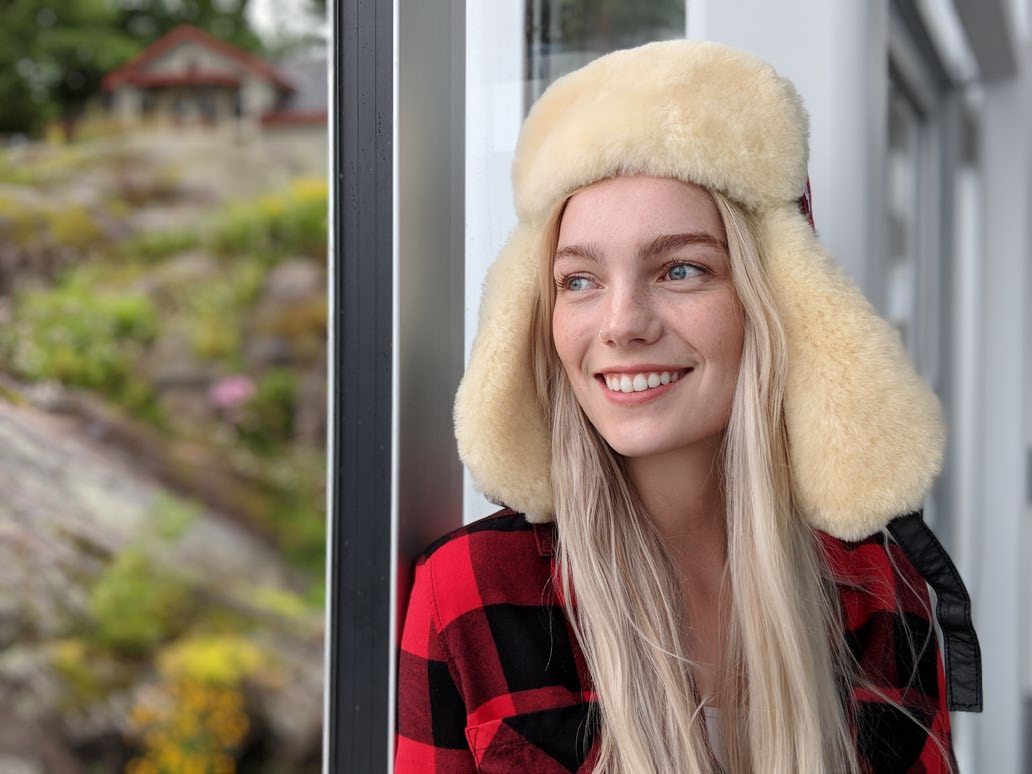}
    Synthetic shallow DoF%
    \end{minipage}
    
    \vspace{1em}
    
     \caption{Shallow depth of field can be computationally introduced by blurring an all-in-focus image using depth estimation and segmentation. Image courtesy of~\citep{wadhwa2018synthetic}.}
    \label{fig:dof-wadhwa2018synthetic}
\end{figure}

The range of the depth of field is inversely proportional to the size of the camera aperture: a wide aperture produces a shallow depth of field while a narrow aperture leads to a wider depth of field. On mobile phone cameras, physical limitations make it impossible to have a wide aperture. This implies that capturing images with a shallow depth of field is virtually impossible.  Although an all-in-focus image retains the most information, for aesthetic and artistic reasons, users may want to have control of the depth of field.

Recently, mobile phone manufacturers have introduced a computational (shallow) depth-of-field effect called a ``synthetic bokeh'' (see Figure~\ref{fig:dof-wadhwa2018synthetic}).  An accurate depth map estimate would enable computationally introducing spatially varying depth blur and simulate the depth-of-field effect. The traditional solution to estimate a depth map is based on stereo vision and requires two cameras. Adding a second camera introduces additional costs and increases power consumption and size. An alternative is to introduce a dedicated depth sensor based on structured light or time-of-flight technologies. However, these tend to be expensive and mainly work indoors, which significantly restricts their use. Accurately estimating a depth map from a single image is a severely ill-posed problem that generally leads to very limited accuracy. 

\citet{wadhwa2018synthetic} introduced a system to synthetically generate a depth-of-field effect on smartphone cameras. The method runs completely on the device and uses only the information from a single camera (rear or front facing).  The key idea is to incorporate a deep neural network to segment out people and faces, and then use the segmentation to adaptively blur the background. Additionally, if available, the system uses dual-pixel information now present in many hardware auto-focus systems. The dual-pixel data provides very small baseline stereo information that allows algorithmic generation of dense depth maps. 

Meanwhile, the front-facing camera is used almost exclusively to take ``selfie'' photos---that is, a close-up of the face and upper half of the photographer's body. A neural network trained for this type of image allows segmenting the main character out from the background. The background is then appropriately blurred to give the idea of depth of field. When using the rear-facing camera, no prior information about the photograph composition can be assumed. Thus, having dense depth information becomes crucial. 

It is worth mentioning that this computational DoF system does not necessarily lead to a physically plausible photograph as would have been taken by a camera with a wider aperture---it merely suggests the right look. For instance, among other algorithm design choices, all pixels belonging to the segmentation mask are assumed to be in focus even if they are at different depths.

\section{The Future and Upcoming Challenges}
Mobile cameras have made significant strides in quality, matching (and even surpassing) the image quality of so-called micro-4/3 standalone cameras\footnote{~https://www.dpreview.com/articles/6476469986/dpreview-products-of-the-year-2018?slide=25}. While it seems unlikely that, absent a major change in form factor, smartphone cameras will equal DSLR quality, much still remains to be done. Here are some promising avenues and new challenges. 

\subsection{Algorithmics}
The impressive success of neural networks in computer vision has not (yet) been widely replicated in practical aspects of computational photography. Indeed, the impressive progress in mobile imaging has largely been facilitated by methods that are mostly {\em not} based on deep neural networks (DNN). Given the proliferation of DNNs in every other aspect of technology, this may seem surprising. Two observations may help explain this landscape. First, DNNs still have relatively heavy computing and memory requirements that are mostly outside the scope of current capabilities of mobile devices. This may change soon. Second, resource limitations aside, DNN-based (particularly the so-called \emph{generative}) models still have the tendency to produce certain artifacts in the final images that are either undesirable, or intolerable in a consumer device. Furthermore, such errors are not easily diagnosed and repaired because, unlike existing methods, ``tuning'' the behavior of deep models is not easy. These issues too will be remedied in due time. Meanwhile, DNN approaches continue to be developed with the intention to replace the entire end-to-end processing pipeline---examples include   DeepISP~\citep{schwartz2018deepisp} and ~\citep{TimofteCVPR2020}, to name just two. 

\subsection{Curation} 
Today nearly everyone who can afford to have a smartphone owns one. And we now take and share more photos than ever. Given how little cost and effort picture-taking entails, we've also evolved a tendency to often capture multiple pictures of the same subject. Yet, typically only a few of our many shots turn out well, or to our liking. As such, storage, curation, and retrieval of photographs have become another aspect of photography that has drawn attention recently and deserves much more work. Some recent methods \citep{Talebi2018NIMA} have developed neural network models trained on many images annotated for technical and aesthetic quality, which now enable machine evaluation of images in both qualitative and quantitative terms. Of course, this technology is in very early stages of development and represents aggregate opinion, not necessarily meant to cater to personal taste yet. Similar models can also rank photos based on their technical quality---aspects such as whether the subject is well lit, centered, and in focus. Needless to say, much work remains to be done here. 
\subsection{Broader use cases}
The proliferation of cameras and computational photography technology is of course not limited to the mobile platform. It is indeed not an exaggeration to say that cameras are nearly \emph{everywhere}. Many of the techniques developed for the mobile platform may in fact be useful for enhancing the quality of images derived on other platforms, notably scientific instrumentation, automotive imaging, satellite imaging, and more. But caution is warranted, and key differences should be noted.  

In particular it is important to note that the imaging pipelines developed in the context of mobile photography are specifically tuned for producing aesthetically pleasing images. Meanwhile, in scientific uses of computational photography, the end goal is not the image itself but rather certain, and varied, information extracted from the images. For instance, in the medical realm the end task maybe diagnostic, and this may not be best facilitated by a ``pretty'' picture. Instead, what is required is a maximally ``informative'' picture. Correspondingly, cameras on mobile devices are not built, configured, or tuned to provide such information \footnote{~For instance, automatic white balance can be counter-productive if the end goal is to make a physical measurement that depends on color fidelity in a different sense.}. An interesting case study is the role that cameras have played (or could better have played) in recent environmentally calamitous events such as the massive wildfires in California. Nearly all cameras, tuned to normal viewing conditions, and biased toward making the pictures pleasing, were largely unable to capture the true physical attributes of the dark, orange-hued skies polluted with smoke\footnote{https://www.theatlantic.com/ideas/archive/2020/11/photography-has-never-known-how-handle-climate-change/617224/}. 

The bottom line is that mobile cameras, properly re-imagined and built, can play an even more useful, helpful, and instrumental role in our lives than they do today.

\subsection{Epilogue} The technology behind computational photography has advanced rapidly in the last decade---the science and engineering techniques that generate high-quality images from small mobile cameras will continue to evolve. But so too will our needs and tastes for the types of devices we are willing to carry around, and the kinds of visual or other experiences we wish to record and share. 

It is hard to predict with any certainty what the mobile devices of the future will look like. But as surely as Ansel Adams would not have seen the mobile phone camera coming, we too may be surprised by both the form, and the vast new uses, of these devices in the next decade.   
\section*{Disclosure Statement}
The authors are not aware of any affiliations, memberships, funding, or financial holdings that might be perceived as affecting the objectivity of this review. 

\section*{Acknowledgments}
The authors wish to acknowledge the computational imaging community of scholars and colleagues---industrial and academic alike---whose work has led to the advances reported in this review. While we could not cite them all, we dedicate this paper to their collective work.

\bibliographystyle{ar-style1}
\bibliography{Biblio-tour}

\end{document}